\documentclass[10pt,twocolumn,letterpaper]{article}
\pdfoutput=1
\usepackage{pdfpages}
\usepackage{cvpr}
\usepackage{times}
\usepackage{xcolor}
\usepackage{epsfig}
\usepackage{graphicx}
\usepackage{amsmath}
\usepackage{amssymb}
\usepackage{algorithm}
\usepackage{algorithmic}
\usepackage{caption}

\usepackage{enumitem}

\usepackage[pagebackref=true,breaklinks=true,letterpaper=true,colorlinks,bookmarks=false]{hyperref}

\cvprfinalcopy 

\def\cvprPaperID{2297} 

\begin{document}
\title{MaskGAN: Towards Diverse and Interactive Facial Image Manipulation}

\author{
Cheng-Han Lee\textsuperscript{1}\,\,\,\,\, 
Ziwei Liu\textsuperscript{2}\,\,\,\,\, 
Lingyun Wu\textsuperscript{1}\,\,\,\,\,
Ping Luo\textsuperscript{3}\\\\
\centerline{
\textsuperscript{1}SenseTime Research\,\,\,\,\,
\textsuperscript{2}The Chinese University of Hong Kong\,\,\,\,\,
\textsuperscript{3}The University of Hong Kong}
}
\twocolumn[{
\renewcommand\twocolumn[1][]{#1}%
\maketitle
\vspace{-40pt}
\begin{center}
  \centering
  \includegraphics[width=1\textwidth]{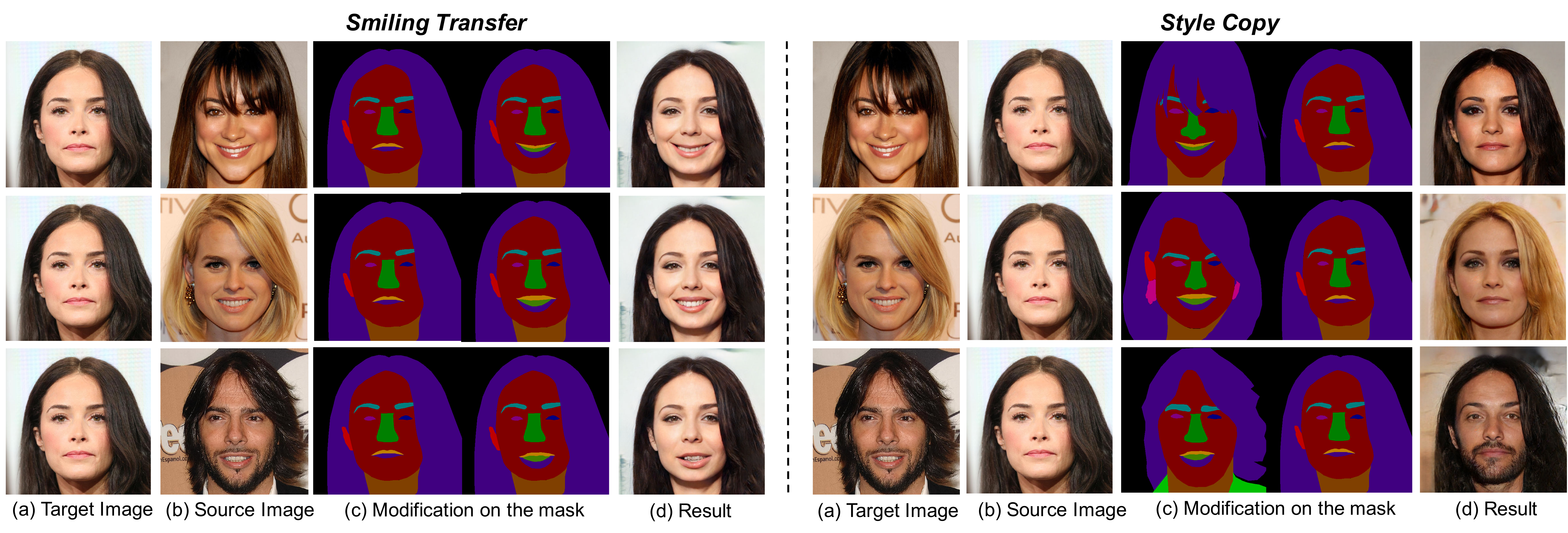}
  \vspace{-20pt}
  \captionof{figure}{\footnotesize Given a target image (a), users are allowed to modify masks of the target images in (c) according to the source images (b) so that we can obtain manipulation results (d). The left shows illustrative examples from ``neutral'' to ``smiling'', while the right shows style copy such as makeup, hair, expression, skin color, etc.}
\end{center}
}]

\begin{abstract}
\vspace{-10pt}
Facial image manipulation has achieved great progress in recent years.
However, previous methods either operate on a predefined set of face attributes or leave users little freedom to interactively manipulate images.
To overcome these drawbacks, we propose a novel framework termed MaskGAN, enabling diverse and interactive face manipulation.
Our key insight is that semantic masks serve as a suitable intermediate representation for flexible face manipulation with fidelity preservation.
MaskGAN has two main components: 1) Dense Mapping Network (DMN) and 2) Editing Behavior Simulated Training (EBST).
Specifically, DMN learns style mapping between a free-form user modified mask and a target image, enabling diverse generation results.
EBST models the user editing behavior on the source mask, making the overall framework more robust to various manipulated inputs. 
Specifically, it introduces dual-editing consistency as the auxiliary supervision signal.
To facilitate extensive studies, we construct a large-scale high-resolution face dataset with fine-grained mask annotations named CelebAMask-HQ.
MaskGAN is comprehensively evaluated on two challenging tasks: attribute transfer and style copy, demonstrating superior performance over other state-of-the-art methods.
The code, models, and dataset are available at \href{https://github.com/switchablenorms/CelebAMask-HQ}{https://github.com/switchablenorms/CelebAMask-HQ}.
\vspace{-10pt}
\end{abstract}

\begin{figure*}[t]
\begin{center}
 \includegraphics[width=0.9\linewidth]{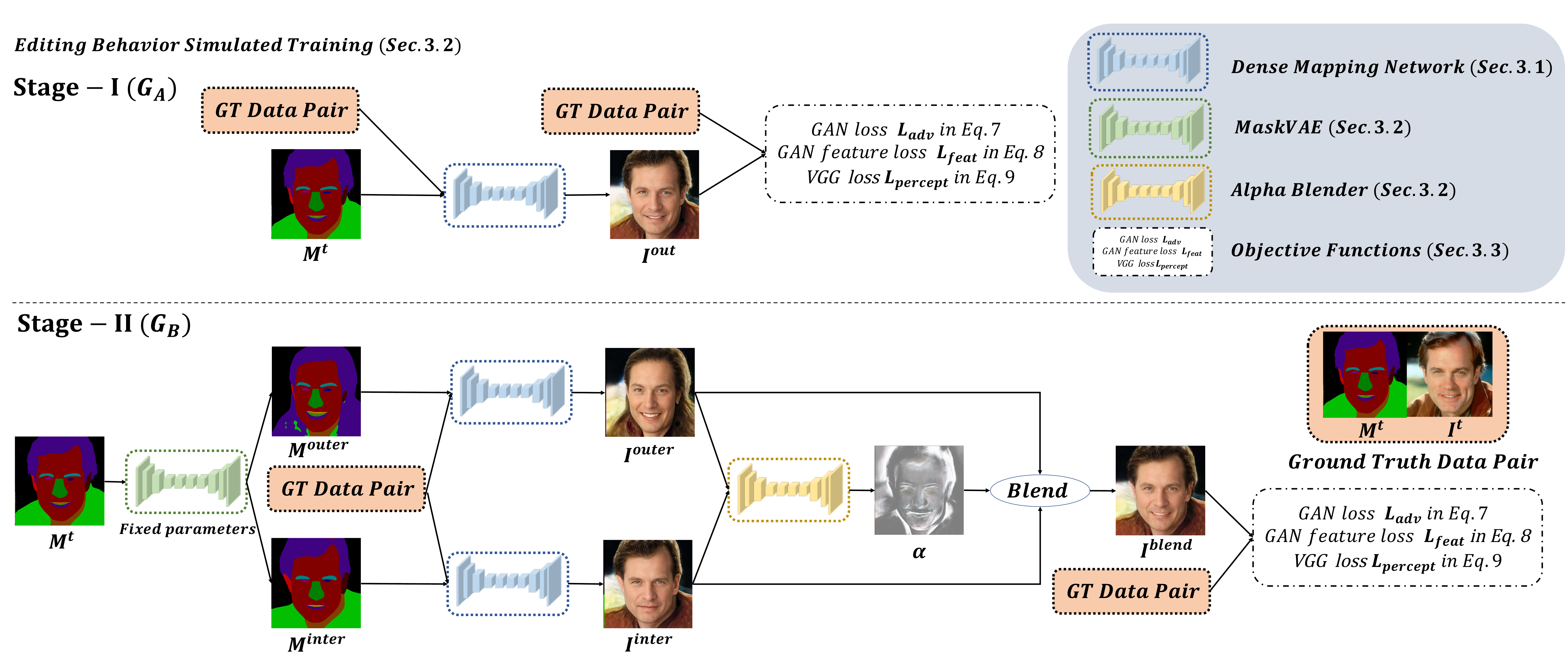}
\end{center}
\vspace{-20pt}
\caption{\footnotesize Overall training pipeline. Editing Behavior Simulated Training can be divided into two stage. After loading the pre-trained model of Dense Mapping Network and MaskVAE, we iteratively update these two stages until model converging.}
\label{fig:long}
\label{fig:onecol}
\label{fig2}
\vspace{-10pt}
\end{figure*} 

\section{Introduction}
Facial image manipulation is an important task in computer vision and computer graphic, enabling lots of  applications such as automatic facial expressions and styles (\eg hairstyle, skin color) transfer.
This task can be roughly categorized into two types: semantic-level manipulation ~\cite{choi2018stargan,liu2017unsupervised,perarnau2016invertible,lample2017fader,li2016deep} and geometry-level manipulation~\cite{yeh2016semantic,xiao2018elegant,yin2018instance,zhang2019one}. 
However, these methods either operate on a pre-defined set of attributes or leave users little freedom to interactively manipulate the face images. 

To overcome the aforementioned drawbacks, we propose a novel framework termed MaskGAN, which aims to enable diverse and interactive face manipulation. 
\emph{Our key insight is that semantic masks serve as a suitable intermediate representation for flexible face manipulation with fidelity preservation.}
Instead of directly transforming images in the pixel space, MaskGAN learns the face manipulation process as traversing on the mask manifold~\cite{liu2015semantic}, thus producing more diverse results with respect to facial components, shapes, and poses.
An additional advantage of MaskGAN is that it provides users an intuitive way to specify the shape, location, and facial component categories for interactive editing.


MaskGAN has two main components including \textbf{1)} Dense Mapping Network and \textbf{2)} Editing Behavior Simulated Training.
The former learns the mapping between the semantic mask and the rendered image, while the latter learns to model the user editing behavior when manipulating masks.
Specifically, Dense Mapping Network consists of an Image Generation Backbone and a Spatial-Aware Style Encoder.
The Spatial-Aware Style Encoder takes both the target image and its corresponding semantic label mask as inputs; it produces spatial-aware style features to the Image Generation Backbone. 
After receiving a source mask with user modification, the Image Generation Backbone learns to synthesize faces according to the spatial-aware style features. 
In this way, our Dense Mapping Network is capable of learning the fine-grained style mapping between a user modified mask and a target image.

Editing behavior simulated training is a training strategy to model the user editing behavior on the source mask, which introduces the dual-editing consistency as the auxiliary supervision signal
Its training pipeline comprises an obtained Dense Mapping Network, a pre-trained MaskVAE, and an alpha blender sub-network.
\emph{The core idea is that the generation results of two locally-perturbed input masks (by traversing on the mask manifold learned by MaskVAE) blending together should retain the subject's appearance and identity information.} 
Specifically, the MaskVAE with encoder-decoder architecture is responsible for modeling the manifold of geometrical structure priors.
The alpha blender sub-network learns to perform alpha blending \cite{porter1984compositing} as image composition, which helps maintain the manipulation consistency.
After training with editing behavior simulation, Dense Mapping Network is more robust to the various changes of the user-input mask during inference.

MaskGAN is comprehensively evaluated on two challenging tasks, including attribute transfer and style copy, showing superior performance compared to other state-of-the-art methods.
To facilitate large-scale studies, we construct a large-scale high-resolution face dataset with fine-grained mask labels named CelebAMask-HQ.
Specifically, CelebAMask-HQ consists of over 30,000 face images of 512$\times$512 resolution, where each image is annotated with a semantic mask of 19 facial component categories, \eg eye region, nose region, mouth region.

To summarize, our contributions are three-fold: \textbf{1)} We present MaskGAN for diverse and interactive face manipulation. Within the MaskGAN framework, Dense Mapping Network is further proposed to provide users an interactive way for manipulating face using its semantic label mask. \textbf{2)} We introduce a novel training strategy termed Editing Behavior Simulated Training, which enhances the robustness of Dense Mapping Network to the shape variations of the user-input mask during inference. \textbf{3)} We contribute CelebAMask-HQ, a large-scale high-resolution face dataset with mask annotations. We believe this geometry-oriented dataset would open new research directions for the face editing and manipulation community.

\section{Related Work}

\noindent\textbf{Generative Adversarial Network.} GAN \cite{goodfellow2014generative} generally consists of a generator and a discriminator that compete with each other. Because GAN can generate realistic images, it enjoys pervasive applications on tasks such as image-to-image translation \cite{isola2017image,zhu2017unpaired,liu2017unsupervised,wang2018high, park2019semantic}, image inpainting \cite{liu2018image, yu2018generative, yu2019free, jo2019sc}, and virtual try-on \cite{hanyangCVPR2020, han2018viton, chou2018pivtons, wang2018toward}.
\noindent\textbf{Semantic-level Face Manipulation.} Deep semantic-level face editing has been studied for a few years. Many works including \cite{choi2018stargan,liu2017unsupervised,perarnau2016invertible,lample2017fader,li2016deep, lee2018attribute} achieved impressive results. 
IcGAN \cite{perarnau2016invertible} introduced an encoder to learn the inverse mappings of conditional GAN.
DIAT \cite{li2016deep} utilized adversarial loss to transfer attributes and learn to blend predicted face and original face. 
Fader Network \cite{lample2017fader} leveraged adversarial training to disentangle attribute related features from the latent space. 
StarGAN \cite{choi2018stargan} was proposed to perform multi-domain image translation using a single network conditioned on the target domain label. However, these methods cannot generate images by exemplars.

\noindent\textbf{Geometry-level Face Manipulation.} Some recent studies~\cite{yeh2016semantic,xiao2018elegant,yin2018instance, gu2019mask} start to discuss the possibility of transferring facial attributes at instance level from exemplars. 
For example, ELEGANT \cite{xiao2018elegant} was proposed to exchange attribute between two faces by exchanging the latent codes of two faces. 
However, ELEGANT \cite{xiao2018elegant} cannot transfer the attributes (e.g. `smiling') from exemplars accurately. 
For 3D-based face manipulation, though 3D-based methods \cite{cao2013facewarehouse, nagano2018pagan, geng2018warp} achieve promising results on normal poses, they are often computationally expensive and their performance may degrade with large and extreme poses.

\section{Our Approach}

\noindent
\textbf{Overall Framework.}
Our goal is to realize structural conditioned face manipulation using MaskGAN, given an target image $I^{t} \in \mathbb{R} ^{H \times W \times 3}$ , a semantic label mask of target image $M^{t} \in \mathbb{R} ^{H \times W \times C}$ and a source semantic label mask $M^{src} \in \mathbb{R} ^{H \times W \times C}$ (user modified mask). 
When users manipulating the structure of $M^{src}$, our model can synthesis a manipulated face $I^{out} \in \mathbb{R} ^{H \times W \times 3}$ where $C$ is the category number of the semantic label.

\noindent
\textbf{Training Pipeline.}
As shown in Fig. \ref{fig2}, MaskGAN composes of three key elements: Dense Mapping Network (DMN), MaskVAE, and Alpha Blender which are trained by Editing Behavior Simulated Training (EBST).
DMN (See Sec. \ref{sec31}) 
provides users an interface for manipulating face toward semantic label mask which can learn a style mapping between $I^{t}$ and $M^{src}$.
MaskVAE is responsible for modeling the manifold of structure priors (See Sec. \ref{sec32}).
Alpha Blender is responsible for maintaining manipulation consistency (See Sec. \ref{sec32}).
To make DMN more robust to the changing of the user-defined mask $M^{src}$ in the inference time, we propose a novel training strategy called EBST (See Sec. \ref{sec32}) which can model the user editing behavior on the $M^{src}$.
This training method needs a well trained DMN, a MaskVAE trained until low reconstruction error, and an Alpha Blender trained from scratch.
The training pipeline can be divided into two stages.
In training stage, we replace $M^{src}$ with $M^{t}$ as input.
In Stage-I, we update DMN with $M^{t}$ and $I^{t}$ firstly.
In Stage-II, we used MaskVAE to generate two new mask $M^{inter}$ and $M^{outer}$ with small different from $M^{t}$ and generate two faces $I^{inter}$ and $I^{outer}$. Then, Alpha Blender blends these two faces to $I^{blend}$ for maintaining manipulation consistency.
After EBST, DMN would be more robust to the change of $M^{src}$ in the inference stage.
The details of the objective functions are shown in Sec. \ref{sec33}.

\noindent
\textbf{Inference Pipeline.}
We only need DMN in testing. In Fig. \ref{fig3}, different from training stage, we simply replace the input of Image Generation Backbone with $M^{src}$ where $M^{src}$ can be defined by the user.


\begin{figure}[t]
\vspace{-10pt}
\begin{center}
 \includegraphics[width=0.9\linewidth]{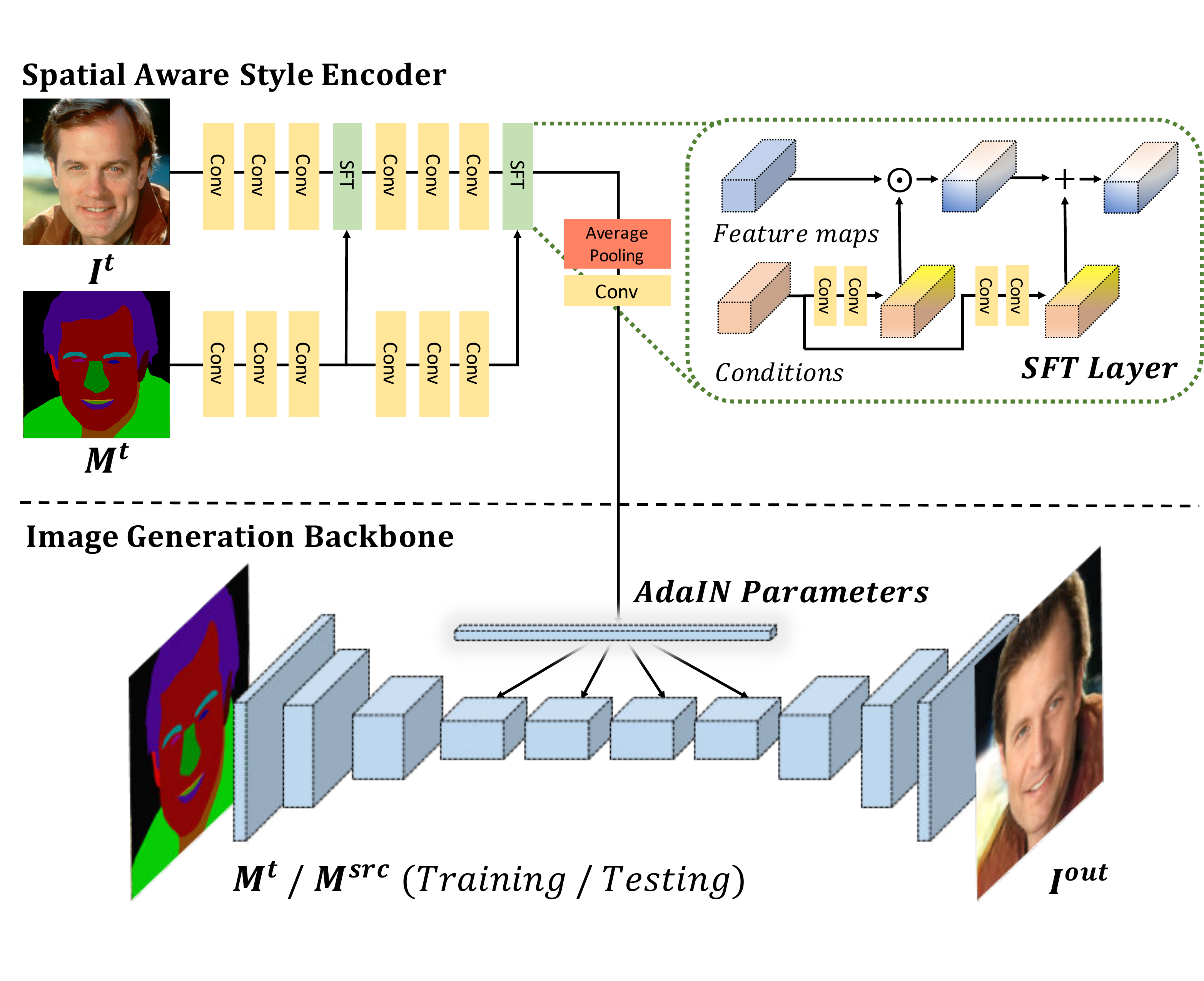}
\end{center}
\vspace{-25pt}
\caption{\footnotesize Architecture of Dense Mapping Network which is composed of a \textbf{Spatial-Aware Style Encoder} and a \textbf{Image Generation Backbone}.}
\label{fig:long}
\label{fig:onecol}
\label{fig3}
\vspace{-10pt}
\end{figure}  
\subsection{Dense Mapping Network}
\label{sec31}
 Dense Mapping Network adopts the architecture of Pix2PixHD as a backbone and we extend it with an external encoder $Enc_{style}$ which will receive $I^{t}$ and $M^{t}$ as inputs. The detailed architecture is shown in Fig. \ref{fig3}. 


\noindent
\textbf{Spatial-Aware Style Encoder.}
We propose a Spatial-Aware Style Encoder network $Enc_{style}$ which receives style information $I^{t}$ and its corresponding spatial information $M^{t}$ at the same time. 
To fuse these two domains, we utilize Spatial Feature Transform (SFT) in SFT-GAN \cite{wang2018recovering}. 
The SFT layer learns a mapping function $\mathcal{M}:\Psi\mapsto(\gamma, \beta)$ where affine transformation parameters $(\gamma, \beta)$ is obtained by prior condition $\Psi$ as $(\gamma, \beta) = \mathcal{M}(\Psi)$. 
After obtaining $\gamma$ and $\beta$, the SFT layer both perform feature-wise and spatial-wise modulation on feature map F as 
 $SFT(F|\gamma, \beta) = \gamma \odot F + \beta$ where the dimension of F is the same as $\gamma$ and $\beta$, and $\odot$ is referred to element-wise product. 
Here we obtain the prior condition $\Psi$ from the features of $M^{t}$ and feature map F from $I^{t}$. Therefore, we can condition spatial information $M^{t}$ on style information $I^{t}$ and generate $x_{i}, y_{i}$ as following:
\begin{equation}
x_{i}, y_{i} = Enc_{style}(I^{t}_i, M^{t}_i) 
\label{eq1},
\end{equation}
where $x_{i}, y_{i}$ are affine parameters which contain spatial-aware style information.
To transfer the spatial-aware style information to target mask input, we leverage adaptive instance normalization \cite{huang2017adain} (AdaIN) on residual blocks $z_{i}$ in the DMN. 
The AdaIN operation which is a state-of-the-art method in style transfer is defined as:
\begin{equation}
AdaIN(z_{i}, x_{i}, y_{i}) = x_{i}(\frac{z_{i} - \mu(z_{i})}{\sigma(z_{i})}) + y_{i}
\label{eq2},
\end{equation}
which is similar to Instance Normalization \cite{ulyanov2016instance}, but replaces the affine parameters from IN with conditional style information.

DMN is a generator defined as $G_{A}$ where
$I^{out} = G_{A}(Enc_{style}(I^{t}, M^{t}),M^{t}))$.
With the Spatial-Aware Style Encoder, DMN learns the style mapping between $I^{t}$ and $M^{src}$ according to the spatial information provided by $M^{t}$. Therefore, styles (e.g. hairstyle and skin style) in $I^{t}$ are transitioned to the corresponding position on $M^{src}$ so that DMN can synthesis final manipulated face $I^{out}$.   

\subsection{Editing Behavior Simulated Training}\label{sec32}
Editing Behavior Simulated Training can model the user editing behavior on the $M^{src}$ in training time. 
This training method needs a well trained Dense Mapping Network $G_{A}$, a MaskVAE trained until low reconstruction error, and an Alpha Blender trained from scratch.
MaskVAE composed of $Enc_{\mathrm{VAE}}$ and $Dec_{\mathrm{VAE}}$, which is responsible for modeling the manifold of structure priors.
Alpha Blender $B$ is responsible for maintaining manipulation consistency.
We define $G_{B}$ as another generator which utilize MaskVAE, DMN, and Alpha Blender as $G_{B}$ where $G_{B} \equiv B(G_{A}(I^{t}, M^{t},M^{inter}), G_{A}(I^{t}, M^{t},M^{outer}))$.
The overall training pipeline is shown in Fig. \ref{fig2} and the detailed algorithm is shown in Algo. \ref{alg1}.
Our training pipeline can be divided into two stages.
Firstly, we need to load pretrained model of $G_{A}$, $Enc_{\mathrm{VAE}}$ and $Dec_{\mathrm{VAE}}$.
In stage-I, we update $G_{A}$ once. 
In stage-II, given $M^{t}$, we obtain two new masks $M^{inter}$ and $M^{outer}$ with small structure interpolation and extrapolation from the original one by adding two parallel vectors with reverse direction on the latent space of the mask.
These vectors are obtained by $\pm \frac{z^{ref} - z^{t}}{\lambda_{inter}}$ where $z^{ref}$ is latent representation of a random selected mask $M^{ref}$ and $\lambda_{inter}$ is set to $2.5$ for appropriate blending.
After generating two faces by DMN, Alpha Blender learns to blend two images toward the target image where keeping the consistency with the original one. Then, we iteratively update the $G_{A}$ and $G_{B}$ ($Stage-I$ and $Stage-II$ in Fig. \ref{fig2}) until model converging. After EBST, DMN would be more robust to the change of the user-modified mask in inference time.

\begin{algorithm}[t]
\footnotesize
  \caption{Editing Behavior Simulated Training}
  \label{alg1}
  {\bf Initialization:}
    Pre-trained $G_{A}, Enc_{\mathrm{VAE}}, Dec_{\mathrm{VAE}}$ models
    
  {\bf Input:}
    $I^{t}, M^{t}, M^{ref}$
    
  {\bf Output:}
    $I^{out}, I^{blend}$
  \begin{algorithmic}[1]
  \WHILE{iteration not converge}
  \STATE Choose one minibatch of $N$ mask and image pairs $\left \{M^{t}_i, M^{ref}_i, I^{t}_i \right \}$, $i=1,...,N$.
  \STATE $z^{t} = Enc_{\mathrm{VAE}}(M^{t})$
  \STATE $z^{ref} = Enc_{\mathrm{VAE}}(M^{ref})$
  \STATE $z^{inter}, z^{outer} = z^{t} \pm \frac{z^{ref} - z^{t}}{\lambda_{inter}}$
  \STATE $M^{inter} = Dec_{\mathrm{VAE}}(z^{inter})$
  \STATE $M^{outer} = Dec_{\mathrm{VAE}}(z^{outer})$
  \STATE Update $G_{A}(I^{t}, M^{t})$ with Eq. \ref{eq6}
  \STATE Update $G_{B}(I^{t}, M^{t}, M^{inter}, M^{outer})$ with Eq. \ref{eq6}
  \ENDWHILE 
  \end{algorithmic}
\end{algorithm}

\begin{figure}[t]
\begin{center}
 \includegraphics[width=0.9\linewidth]{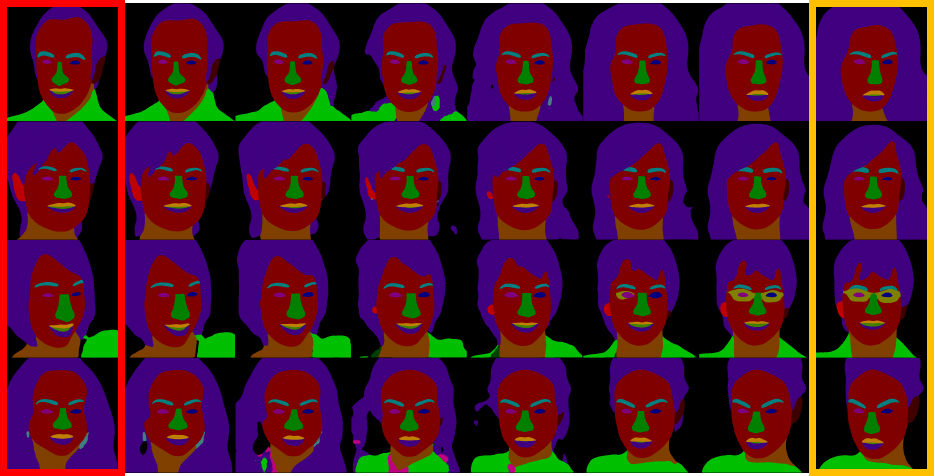}
\end{center}
\vspace{-20pt}
\caption{\footnotesize Samples of linear interpolation between two masks (between the red block and  the orange block). MaskVAE can perform smooth transition on masks.}
\label{fig:long}
\label{fig:onecol}
\label{fig4}
\vspace{-10pt}
\end{figure}  

\noindent\textbf{Structural Priors by MaskVAE.}
Similar to Variational Autoencoder \cite{kingma2013auto}, the objective function for learning a MaskVAE consists of two parts: (i) ${L}_{reconstruct}$, which controls the pixel-wise semantic label difference, (ii) ${L}_{KL}$, which controls the smoothness in the latent space. 
The overall objective is to minimize the following loss function:
\begin{equation}
\mathcal{L}_{MaskVAE} = \mathcal{L}_{reconstruct} + \lambda_{KL}\mathcal{L}_{KL} 
\label{eq3},
\end{equation}
where $\lambda_{KL}$ is set to $1e^{-5}$ which is obtained through cross validation.
The encoder network $Enc_{\mathrm{VAE}}(M^{t})$ outputs the mean $\mu$ and covariance $\sigma$ of the latent vector.
We use KL divergence loss to minimize the gap between the prior $P(z)$ and the learned distribution, $i.e.$
\begin{equation}
\mathcal{L}_{KL} = \frac{1}{2} (\mu\mu^{T} + \sum^{J}_{j-1}(exp(\sigma)-\sigma-1))
\label{eq4},
\end{equation}
where denotes the $j-th$ element of vector $\sigma$.
Then, we can sample latent vector by $z = \mu + r \odot exp(\sigma)$ in the training phase, where $r \sim N(0, I)$ is a random vector and $\odot$ denotes element-wise multiplication.

The decoder network $Dec_{\mathrm{VAE}}(z)$ outputs the reconstruct semantic label and calculates pixel-wise cross-entropy loss as follow:
\begin{equation}
\mathcal{L}_{reconstruct} = - \mathbb{E}_{z \sim P(z)}[log (P(M^{t}|z))]
\label{eq5}.
\end{equation}

Fig. \ref{fig4} shows samples of linear interpolation between two masks. MaskVAE can perform smooth transition on masks and EBST relies on a smooth latent space to operate.

\noindent\textbf{Manipulation Consistency by Alpha Blender.}
To maintain the consistency of manipulation between $I^{blend}$ and $I^{t}$, we realize alpha blending \cite{porter1984compositing} used in image composition by a deep neural network based Alpha Blender $B$ which learn the alpha blending weight $\alpha$ with two input images : $I^{inter}$ and $I^{outer}$ as $\alpha = B(I^{inter}, I^{outer})$. 
After learning appropriated $\alpha$, Alpha Blender blend $I^{inter}$ and $I^{outer}$ according $I^{blend} = \alpha \times I^{inter} + (1-\alpha) \times I^{outer}$.
As shown in the $Stage-II$ of Fig. \ref{fig2}, Alpha Blender is jointly optimized with two share weighted Dense Mapping Networks. The group of models is defined as $G_{B}$.

\begin{figure}[t]
\begin{center}
 \includegraphics[width=0.8\linewidth]{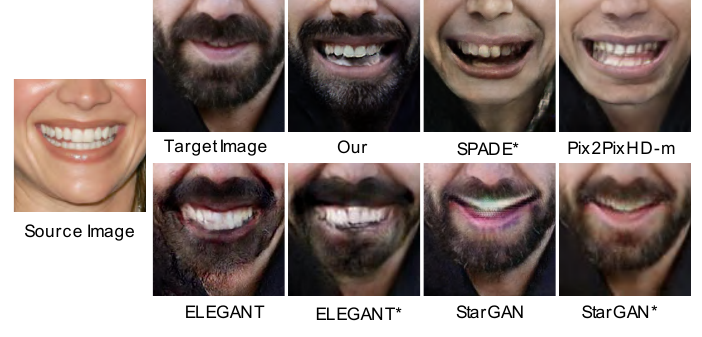}
\end{center}
\vspace{-20pt}
\caption{\footnotesize Zoom in the performance for a specific attribute: \textbf{Smiling} on facial attribute transfer. * indicates the model is trained by images with a size of 256 $\times$ 256. Both SPADE \cite{park2019semantic} and Pix2PixHD-m \cite{wang2018high} cannot preserve attributes (e.g. beard) correctly. Besides, ELEGANT \cite{xiao2018elegant} has poor performance on transferring Smiling from the source image with the mouth very opening. Also, StarGAN \cite{choi2018stargan} has limited performance when training on large images (e.g. 512 $\times$ 512).}
\label{fig:long}
\label{fig:onecol}
\label{fig6}
\vspace{-10pt}
\end{figure}  

\subsection{Multi-Objective Learning}\label{sec33}
The objective function for learning both $G_{A}$ and $G_{B}$ consists of three parts: (i) ${L}_{adv}$, which is the conditional adversarial loss that makes generated images more realistic and corrects the generation structure according to the conditional mask $M^{t}$, (ii) ${L}_{feat}$, which encourages generator to produce natural statistic at multiple scales, (iii) ${L}_{percept}$, which improves content generation from low-frequency to high-frequency details in perceptually toward deep features in VGG-19 \cite{simonyan2014very} trained by ImageNet \cite{deng2009imagenet}. To improve the synthesis quality of a high-resolution image, we leverage multi-scale discriminator \cite{wang2018high} to increase the receptive field and decrease repeated patterns appearing in the generated image. We used two discriminators which refer to $D_{1, 2}$ with identical network structure to operate at two different scales. 
The overall objective is to minimize the following loss function.
\begin{equation}
\begin{split}
\mathcal{L}_{G_{A}, G_{B}} = \mathcal{L}_{adv}(G, D_{1, 2})
\\+\lambda_{feat}\mathcal{L}_{feat}(G,D_{1,2})
\\+\lambda_{percept}\mathcal{L}_{percept}(G)
\label{eq6},
\end{split}
\end{equation}
where $\lambda_{feat}$ and $\lambda_{percept}$ are set to $10$ which are obtained through cross validation.

$\mathcal{L}_{adv}$ is the conditional adversarial loss defined by
\begin{equation}
\begin{split}
\mathcal{L}_{adv} = \mathbb{E}[log(D_{1, 2}(I^{t}, M^{t}))] + \mathbb{E}[1-log(D_{1, 2}(I^{out}, M^{t})]
\end{split}
\label{eq7}.
\end{equation}

$\mathcal{L}_{feat}$ is the feature matching loss \cite{wang2018high} which computes the $L1$ distance between the real and generated image using the intermediate features from discriminator by  
\begin{equation}
\mathcal{L}_{feat} = \mathbb{E}\sum_{i=1}\|D_{1, 2}^{(i)}(I^{t}, M^{t}) - D_{1, 2}^{(i)}(I^{out}, M^{t})\|_{1}
\label{eq8}.
\end{equation}

$\mathcal{L}_{percept}$ is the perceptual loss \cite{johnson2016perceptual} which computes the $L1$ distance between the real and generated image using the intermediate features from a fixed VGG-19 \cite{simonyan2014very} model by
\begin{equation}
\mathcal{L}_{percept} = \sum_{i=1}\frac{1}{M_{i}}[\|\phi^{(i)}(I^{t}) - \phi^{(i)}(I^{out})\|_{1}]
\label{eq9}.
\end{equation}

\begin{figure}[t]
\begin{center}
 \includegraphics[width=1\linewidth]{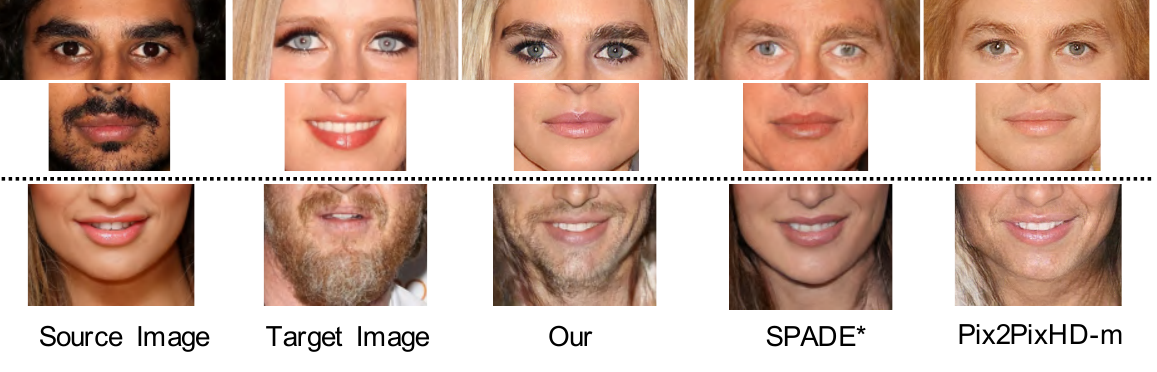}
\end{center}
\vspace{-20pt}
\caption{\footnotesize Zoom in the performance of style copy. Both SPADE \cite{park2019semantic} and Pix2PixHD-m \cite{wang2018high} cannot preserve the attributes -  \textbf{heavy makeup} and \textbf{beard} accurately.}
\label{fig:long}
\label{fig:onecol}
\label{fig8}
\vspace{-10pt}
\end{figure}  

\begin{table}
\caption{\footnotesize Dataset statistics comparisons with an existing dataset. CelebAMask-HQ has superior scales on the number of images and also category annotations.} 
\vspace{-20pt}
\small
\begin{center}
\begin{tabular}{c|c|c} 
\hline
 & Helen \cite{le2012interactive} & CelebAMask-HQ \\ 
\hline
\hline
\# of Images &  2.33K &  \textbf{30K} \\
\hline
Mask size & 400 $\times$ 600 & \textbf{512 $\times$ 512} \\
\hline
\# of Categories & 11 & \textbf{19} \\
\hline
\end{tabular}
\end{center}
\vspace{-10pt}
\label{tab1}
\end{table}

\begin{table*}
\caption{\footnotesize Evaluation on geometry-level facial attribute transfer. Quantitative comparison with other methods for the specific attribute - \textbf{Smiling}. * indicates the model is trained by images with a size of 256 $\times$ 256. $\dagger$ indicates the model is trained with \textbf{Editing Behavior Simulated Training}. StarGAN and ELEGANT have better FID scores, but lower attribute classification accuracy. Pix2PixHD-m obtains the best classification accuracy but has inferior FID scores than others. Although MaskGAN cannot achieve the best FID score, it has relatively higher classification accuracy and segmentation accuracy.} 
\vspace{-12pt}
\footnotesize
\begin{center}
\begin{tabular}{ c|c|c|c|c} 
\hline
Metric & Attribute cls. accuracy(\%) & Segmentation(\%) & FID score & Human eval.(\%)  \\ 
\hline
\hline
StarGAN* \cite{choi2018stargan} & \textcolor{blue}{92.5} & - & 40.61 & - \\ 
StarGAN \cite{choi2018stargan} & 88.0 & - & \textcolor{blue}{30.17} & 7 \\ 
ELEGANT* \cite{xiao2018elegant} & 72.8 & - & 55.43 & - \\
ELEGANT \cite{xiao2018elegant} & 66.5 & - & 35.89 & 34 \\
Pix2PixHD-m \cite{wang2018high} & 78.5 & 93.82 & 54.68 & 13 \\
SPADE* \cite{park2019semantic} & 73.8 & \textcolor{blue}{94.11} & 56.21 & 5 \\
\hline
MaskGAN & 72.3 & 93.23 & \textcolor{red}{46.67} & - \\ 
MaskGAN$^{\dagger}$ & \textcolor{red}{77.3} & \textcolor{red}{93.86} & 46.84 & \textcolor{red}{41} \\ 
\hline
GT & 92.3 & 92.11 & - & - \\
\hline
\end{tabular}
\end{center}
\vspace{-14pt}
\label{tab2}
\end{table*}

\begin{figure*}[t]
\begin{center}
 \includegraphics[width=0.9\linewidth]{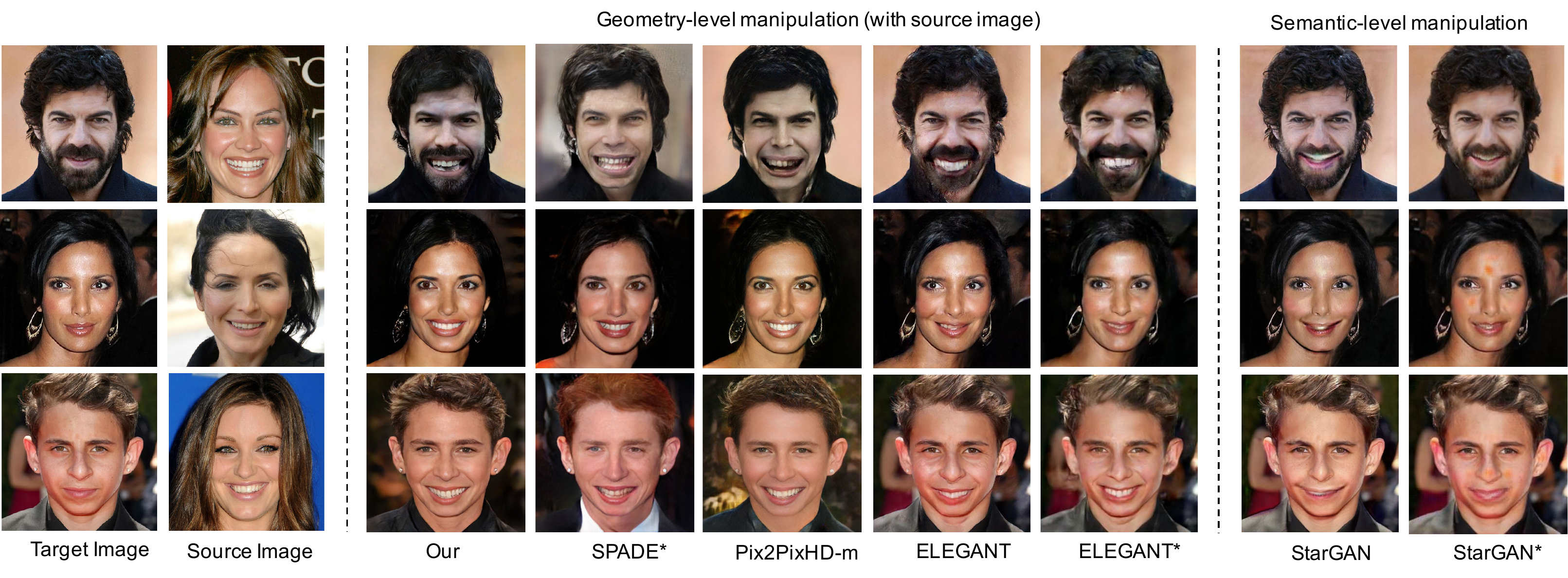}
\end{center}
   \vspace{-20pt}
   \caption{\footnotesize Visual comparison with other methods for a specific attribute: \textbf{Smiling} on facial attribute transfer. * means the model is trained by images with a size of 256 $\times$ 256. The first two columns are target and source pairs. The middle five columns show the results of geometry-level manipulation (our MaskGAN, SPADE \cite{park2019semantic}, Pix2PixHD-m \cite{wang2018high}, and ELEGANT \cite{xiao2018elegant}) which utilize source images as exemplars. The last two columns show the results based on semantic-level manipulation (e.g. StarGAN \cite{choi2018stargan}). StarGAN fails in the region of smiling. ELEGANT has plausible results but sometimes cannot transfer smiling from the source image accurately. Pix2PixHD-m has lower perceptual quality than others. SPADE has poor attribute keeping ability. Our MaskGAN has plausible visual quality and relatively better geometry-level smiling transferring ability.}
\label{fig:long}
\label{fig:onecol}
\label{fig5}
\vspace{-10pt}
\end{figure*}

\section{CelebAMask-HQ Dataset}
We built a large-scale face semantic label dataset named CelebAMask-HQ, which was labeled according to CelebA-HQ \cite{karras2017progressive} that contains 30,000 high-resolution face images from CelebA~\cite{Liu_2015_ICCV}. It has several appealing properties:
\begin{itemize}[leftmargin=*]
\item \textbf{Comprehensive Annotations.} CelebAMask-HQ was precisely hand-annotated with the size of 512 $\times$ 512 and 19 classes including all facial components and accessories such as `skin', `nose', `eyes', `eyebrows', `ears', `mouth', `lip', `hair', `hat', `eyeglass', `earring', `necklace', `neck', and `cloth'.
\item \textbf{Label Size Selection.} The size of images in CelebA-HQ \cite{karras2017progressive} were 1024 $\times$ 1024. However, we chose the size of 512 $\times$ 512 because the cost of the labeling would be quite high for labeling the face at 1024 $\times$ 1024. Besides, we could easily extend the labels from 512 $\times$ 512 to 1024 $\times$ 1024 by nearest-neighbor interpolation without introducing noticeable artifacts.
\item \textbf{Quality Control.} After manual labeling, we had a quality control check on every single segmentation mask. Furthermore, we asked annotaters to refine all masks with several rounds of iterations.
\item \textbf{Amodal Handling.} For occlusion handling, if the facial component was partly occluded, we asked annotators to label the occluded parts of the components by human inferring. On the other hand, we skipped the annotations for those components that are totally occluded.
\end{itemize}
Table \ref{tab1} compares the dataset statistics of CelebAMask-HQ with Helen dataset \cite{le2012interactive}. 

\section{Experiments}

We comprehensively evaluated our approach by showing quantitative and visual quality on different benchmarks.

\subsection{Datasets}
\noindent\textbf{CelebA-HQ.}
\cite{karras2017progressive} is a high quality facial image dataset that consists of 30000 images picked from CelebA dataset \cite{Liu_2015_ICCV}. 
These images are processed with quality improvement to the size of 1024$\times$1024. We resize all images to the size of 512$\times$512 for our experiments.

\noindent\textbf{CelebAMask-HQ.}
Based on CelebA-HQ, we propose a new dataset named CelebAMask-HQ which has 30000 semantic segmentation labels with a size of 512 $\times$ 512.  
Each label in the dataset has 19 classes.

\subsection{Implementation Details}

\noindent\textbf{Network Architectures.}
Image Generation Backbone in Dense Mapping Network follows the design of Pix2PixHD \cite{wang2018high} with 4 residual blocks. Alpha Blender also follows the design of Pix2PixHD but only downsampling 3 times and using 3 residual blocks. The architecture of MaskVAE is similar to UNet \cite{ronneberger2015u} without skip-connection. Spatial-Aware Style Encoder in DMN does not use any Instance Normalization \cite{ulyanov2016instance} layers which will remove style information. All the other convolutional layers in DMN, Alpha Blender, and Discriminator are followed by IN layers. MaskVAE utilizes Batch Normalization \cite{ioffe2015batch} in all layers. 

\noindent\textbf{Comparison Methods.}
We choose state-of-the-art StarGAN \cite{choi2018stargan}, ELEGANT \cite{xiao2018elegant}, Pix2PixHD \cite{wang2018high}, SPADE \cite{park2019semantic} as our baselines. StarGAN performs semantic-level facial attribute manipulation. ELEGANT performs geometry-level facial attribute manipulation. Pix2PixHD performs photo-realistic image synthesis from the semantic mask. We simply remove the branch for receiving $M^{t}$ in Spatial-Aware Style Encoder of Dense Mapping Network as a baseline called Pix2PixHD-m. SPADE performs structure-conditional image manipulation on natural images.

\begin{table*}
\caption{\footnotesize Evaluation on geometry-level style copy. Quantitative comparison with other methods. $\dagger$ indicates the model is trained with \textbf{Editing Behavior Simulated Training}. * indicates the model is trained by images with a size of 256 $\times$ 256. Attribute types in attribute classification accuracy from left to right are \textbf{Male}, \textbf{Heavy Makeup}, and \textbf{No Beard}. MaskGAN has relatively high attribute classification accuracy than Pix2PixHD-m. \textbf{Editing Behavior Simulated Training} further improves the robustness of attribute keeping ability so that MaskGAN$^{\dagger}$ has higher attribute classification accuracy and human evaluation score than MaskGAN.}
\vspace{-12pt}
\footnotesize
\begin{center}
\begin{tabular}{c|c|c|c|c} 
\hline
Metric & Attribute cls. accuracy(\%) & Segmentation(\%) & FID score & Human eval.(\%)  \\ 
\hline
\hline
Pix2PixHD-m \cite{wang2018high} & \textcolor{blue}{56.6}\hspace{0.5cm}\textcolor{blue}{55.1}\hspace{0.5cm}\textcolor{blue}{78.9} & 91.46 & \textcolor{blue}{39.65} & 18 \\
SPADE* \cite{park2019semantic} &
54.5\hspace{0.5cm}51.0\hspace{0.5cm}71.9 & \textcolor{blue}{94.60} & 46.17 & 10 \\
\hline
MaskGAN & 68.1\hspace{0.5cm}72.1\hspace{0.5cm}88.4 & \textcolor{red}{92.34} & 37.55 & 28 \\ 
MaskGAN$^{\dagger}$ &\textcolor{red}{71.7}\hspace{0.5cm}\textcolor{red}{73.3}\hspace{0.5cm}\textcolor{red}{89.5}& 92.31 & \textcolor{red}{37.14}&\textcolor{red}{44} \\ 
\hline
GT & 96.1\hspace{0.5cm}88.5\hspace{0.5cm}95.1 & 92.71 & - & - \\ 
\hline
\end{tabular}
\end{center}
\label{tab3}
\vspace{-10pt}
\end{table*}

\begin{figure*}[t]
\begin{center}
 \includegraphics[width=0.9\linewidth]{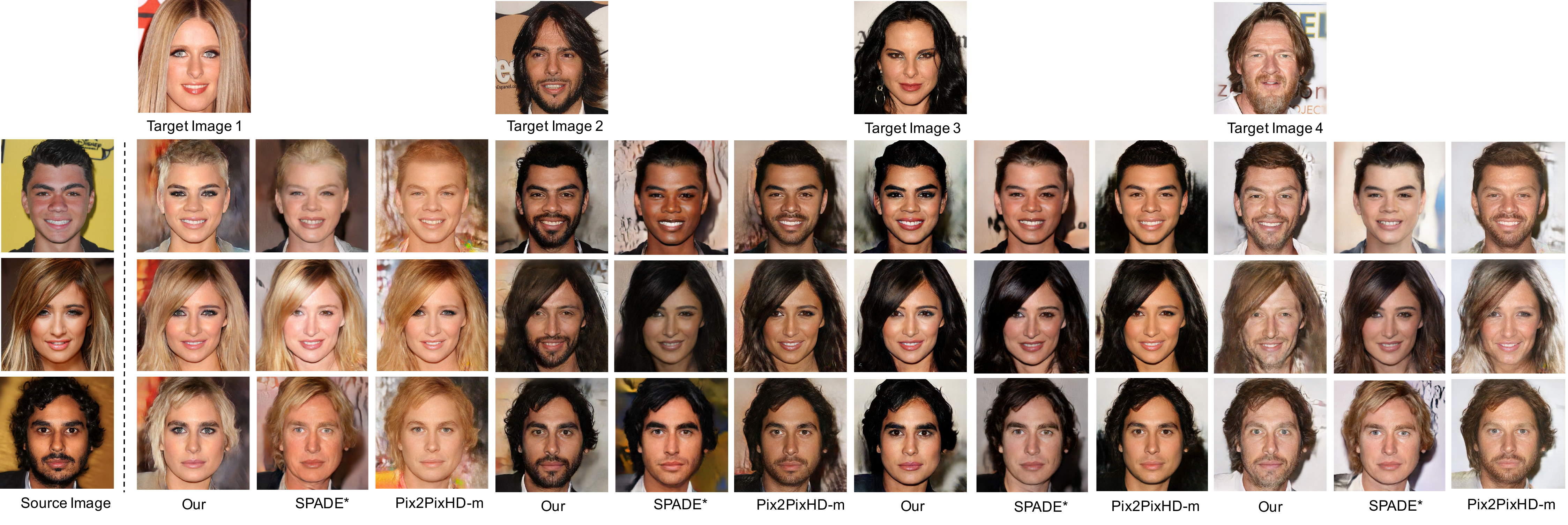}
\end{center}
   \vspace{-20pt}
   \caption{\footnotesize Visual comparison with other methods on style copy. * indicates the model is trained by images with a size of 256 $\times$ 256. All the columns show the results of the proposed method, SPADE \cite{park2019semantic} and Pix2PixHD-m \cite{wang2018high} for four different target images. MaskGAN shows a better ability to transfer style like makeup and gender than SPADE and Pix2PixHD-m. SPADE gets better accuracy on segmentation results.}
\label{fig:long}
\label{fig:onecol}
\label{fig7}
\vspace{-10pt}
\end{figure*}

\subsection{Evaluation Metrics}
\noindent\textbf{Semantic-level Evaluation.}
To evaluate a method of manipulating a target attribute, we examined the classification accuracy of synthesized images. We trained binary facial attribute classifiers for specific attributes on the CelebA dataset by using ResNet-18 \cite{he2016deep} architecture.

\noindent\textbf{Geometry-level Evaluation.}
To measure the quality of mask-conditional image generation, we applied a pre-trained a face parsing model with U-Net \cite{ronneberger2015u} architecture to the generated images and measure the consistency between the input layout and the predicted parsing results in terms of pixel-wise accuracy.  

\noindent\textbf{Distribution-level Evaluation.}
To measure the quality of generated images from different models, we used the Fréchet Inception Distance \cite{heusel2017gans} (FID) to measure the quality and diversity of generated images.

\noindent\textbf{Human Perception Evaluation.}
We performed a user survey to evaluate perceptual generation quality. Given a target image (and a source image in the experiment of style copy), the user was required to choose the best-generated image based on two criteria: 1) quality of transfer in attributes and style 2) perceptual realism. The options were randomly shuffled images generated from different methods.

\noindent\textbf{Identity Preserving Evaluation.}
To further evaluate the identity preservation ability, we conducted an additional face verification experiment by ArcFace \cite{deng2019arcface} (99.52\% on LFW). In the experimental setting, we selected 400 pairs of faces from testing set in CelebA-HQ, and each pair contained a modified face (Smiling) and an unmodified face. Besides, in the testing stage, each face was resized to 112 $\times$ 112.

\subsection{Comparisons with Prior Works}\label{sec44}
The comparison is performed \wrt three aspects, including semantic-level evaluation, geometry-level evaluation, and distributed-level evaluation. We denote our approach as MaskGAN and MaskGAN$^{\dagger}$ for reference, where $^{\dagger}$ indicates the model is equipped with Editing Behavior Simulated Training. For Pix2PixHD \cite{wang2018high} with modification, we name it as Pix2PixHD-m for reference. 

\noindent\textbf{Evaluation on Attribute Transfer.}
We choose \textbf{Smiling} to compare which is the most challenging attribute type to transfer in previous works. 
To be more specific, smiling would influence the whole expressing of a face and smiling has large geometry variety.
To generate the user-modified mask as input, we conducted head pose estimation on the testing set by using the HopeNet \cite{Ruiz_2018_CVPR_Workshops}. 
With the angle information of roll, pitch, and yaw, we selected 400 source and target pairs with a similar pose from the testing set. 
Then, we directly replaced the mask of mouth, upper lip and lower lip from target mask to source mask.
Fig. \ref{fig5}, Fig. \ref{fig6} and Table \ref{tab2} show the visual results and quantitative results on MaskGAN and state-of-the-art. For a fair comparison, StarGAN* and ELEGANT* mean model trained by images with a size of 256 $\times$ 256. StarGAN has the best classification accuracy and FID scores but fails in the region of smiling for the reason that the performance of StarGAN may be influenced by the size of the training data and network design. ELEGANT has plausible results but sometimes cannot transfer smiling from the source image accurately because it exchanges attributes from source image in latent space. SPADE gets the best segmentation accuracy but has an inferior reconstruction ability than others. As long as the target image does not have spatial information to learn a better mapping with the user-defined mask. MaskGAN has plausible visual quality and relative high classification accuracy and segmentation accuracy.

\noindent\textbf{Evaluation on Style Copy.}
To illustrate the robustness of our model, we test MaskGAN on a more difficult task: geometry-level style copy.
Style copy can also be seen as manipulating a face structure to another face.
We selected 1000 target images from the testing set and the source images were selected from the target images with a different order. For this setting, about half of the pairs are a different gender. 
Fig. \ref{fig7}, Fig. \ref{fig8} and Table \ref{tab3} show the visual results and quantitative results on MaskGAN and state-of-the-art. From the visual results and attribute classification accuracy (from left to right: \textbf{Male}, \textbf{Heavy Makeup}, and \textbf{No Beard}), SPADE obtains the best accuracy on segmentation by using Spatially-Adaptive Normalization, but it fails on keeping attributes (e.g. gender and beard). MaskGAN shows better ability to transfer style like makeup and gender than SPADE and Pix2PixHD-m since it introduces spatial information to the style features and simulates the user editing behavior via dual-editing consistency during training.

\noindent\textbf{Evaluation on identity preserving.}
As the experimental results shown in Table \ref{tab4}, our MaskGAN is superior to other state-of-the-art mask-to-image methods for identity preserving. Actually, we have explored adding face identification loss. However, the performance gain is limited. Therefore, we removed the loss in our final framework.

\begin{table}
\caption{\footnotesize Evaluation on identity preserving. Quantitative comparison with other methods. * indicates the model is trained by images with a size of 256 $\times$ 256. MaskGAN is superior to other state-of-the-art mask-to-image methods for identity preserving.}
\vspace{-12pt}
\footnotesize
\begin{center}
\begin{tabular}{c|c} 
\hline
Metric & Face verification accuracy(\%)  \\ 
\hline
\hline
Pix2PixHD-m \cite{wang2018high} & 58.46\\
SPADE* \cite{park2019semantic} & \textcolor{blue}{70.77}\\
\hline
MaskGAN$^{\dagger}$ & \textcolor{red}{76.41} \\ 
\hline
\end{tabular}
\end{center}
\label{tab4}
\vspace{-10pt}
\end{table}

\subsection{Ablation Study}\label{sec45}
In the ablation study, we consider two variants of our model: (i) MaskGAN and (ii) MaskGAN$^{\dagger}$.

\noindent\textbf{Dense Mapping Network.}
In Fig. \ref{fig6}, we observe that Pix2PixHD-m is influenced by the prior information contained in the user-modified mask. For example, if the user modifies the mask to be a female while the target image looks like a male, the predicted image tends to a female with makeup and no beard. Besides, Pix2PixHD-m cannot transition the style from the target image to the user-modified mask accurately. With Spatial-Aware Style Encoder, MaskGAN not only prevents generated results influenced by prior knowledge in the user-modified mask, but also accurately transfers the style of the target image.

\noindent\textbf{Editing Behavior Simulated Training.}
Table \ref{tab2} and Table \ref{tab3} show that simulating editing behavior in training can prevent content generation in the inference stage from being influenced by structure changing on the user-modified mask. It improves the robustness of attribute keeping ability so that MaskGAN demonstrates better evaluation scores.

\begin{figure}
\begin{center}
 \includegraphics[width=1.0\linewidth]{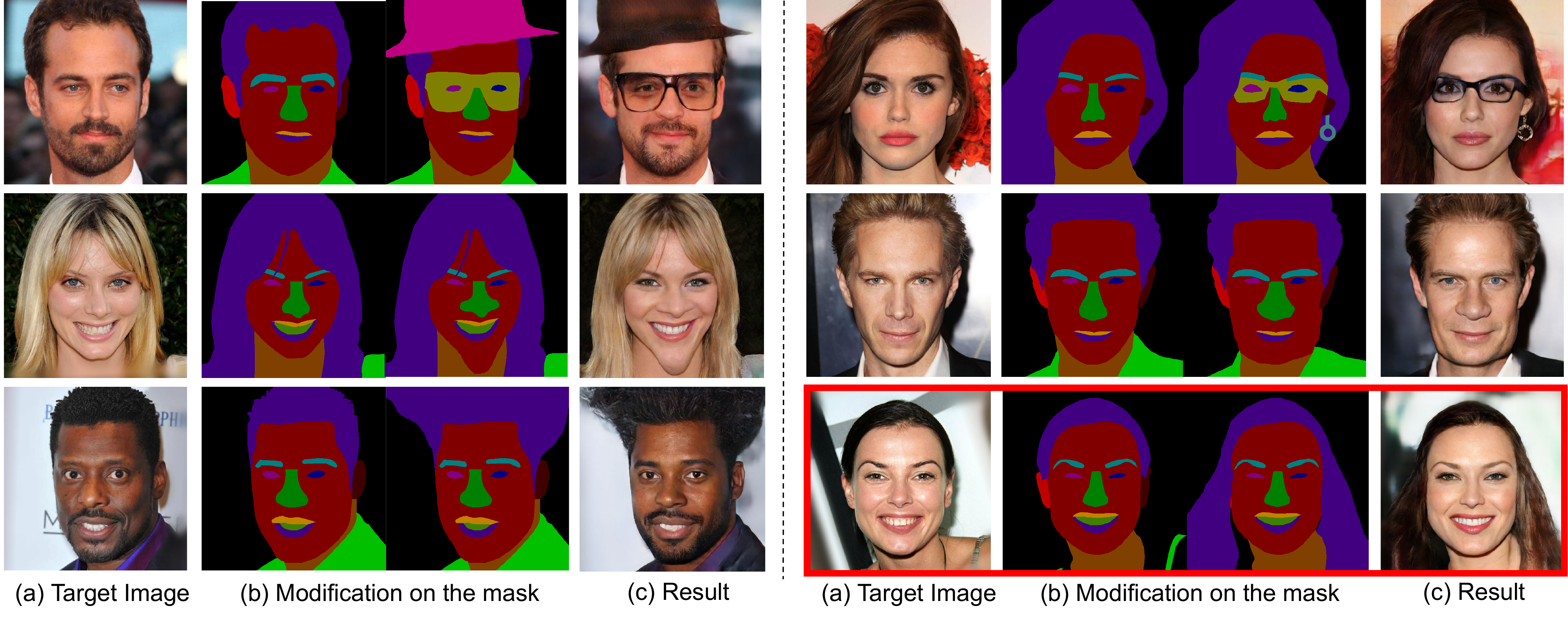}
\end{center}
\vspace{-16pt}
   \caption{\footnotesize Visual results of interactive face editing. The first row shows examples of adding accessories like eyeglasses. The second row shows examples of editing the shape of face and nose. The third row shows examples of adding hair. The red block shows a fail case where the strength of hair color decreases when adding hair to a short hair woman.}
\label{fig:long}
\label{fig:onecol}
\label{fig9}
\vspace{-16pt}
\end{figure}

\subsection{Interactive Face Editing}

%
Our MaskGAN allows users to interactively edit the shape, location, and category of facial components at geometry-level through a semantic mask interface. The interactive face editing results are illustrated in Fig. \ref{fig7}. The first row shows examples of adding accessories like eyeglasses, earrings, and hats. The second row shows examples of editing face shape and nose shape. The third row shows examples of adding hair. More results are in the \textbf{supplementary materials}.

\section{Conclusions}


In this work, we have proposed a novel geometry-oriented face manipulation framework, MaskGAN, with two carefully designed components: 1) Dense Mapping Network and 2) Editing Behavior Simulated Training.
Our key insight is that semantic masks serve as a suitable intermediate representation for flexible face manipulation with fidelity preservation.
MaskGAN is comprehensively evaluated on two challenging tasks: attribute transfer and style copy, showing superior performance over other state-of-the-art methods.
We further contribute a large-scale high-resolution face dataset with fine-grained mask annotations, named CelebAMask-HQ. 
Future work includes combining MaskGAN with image completion techniques to further preserve details on the regions without editing. \\
\textbf{Acknowledgement.} This work was partially supported by HKU Seed Fund for Basic Research, Start-up Fund and Research Donation from SenseTime.
{\small
\bibliographystyle{ieee_fullname}
\bibliography{egbib}
}
\renewcommand\thesection{\Alph{section}}
\setcounter{section}{0}
\newpage\section{Additional Implementation Details}
Our MaskGAN is composed of four key components: MaskVAE, Dense Mapping Network, Alpha Blender, and Discriminator. Specifically, Dense Mapping Network contains two elements: Image Generation Backbone, Spatial-Aware Style Encoder. More details about the architecture design of these components and training details are shown below.

\begin{figure}[t]
\begin{center}
 \includegraphics[width=0.8\linewidth]{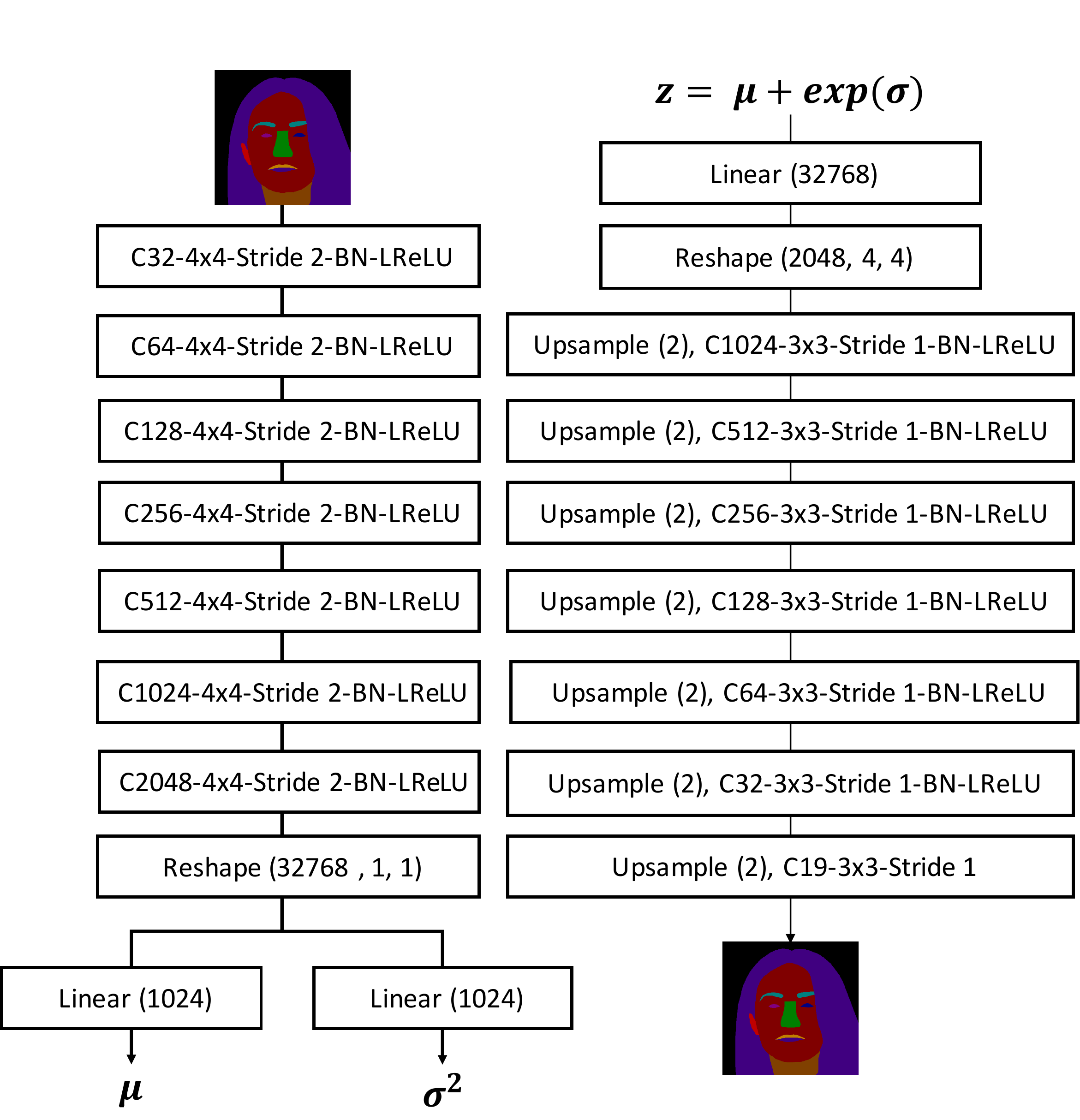}
\end{center}
   \vspace{-10pt}
   \caption{Architecture of MaskVAE.}
\label{fig:long}
\label{fig:onecol}
\label{fig1}
\end{figure}

\noindent
\textbf{MaskVAE.}
The architecture of MaskVAE is similar to UNet \cite{ronneberger2015u} without skip-connection. Detailed architectures of $Enc_{\mathrm{VAE}}$ and $Dec_{\mathrm{VAE}}$ are shown in Fig. \ref{fig1} which uses BN for all layers.

\noindent
\textbf{Image Generation Backbone.} We choose the architecture of Pix2PixHD \cite{wang2018high} as Image Generation Backbone. The detailed architecture is as follow: \\ $c7s1-64,d128,d256,d512,d1024,R1024,R1024,R1024,\\R1024,u512,u256,u128,u64-c7s1$. \\ We utilize AdaIN \cite{huang2017adain} for all residual blocks, other layers use IN.
We do not further utilize a local enhancer because we conduct all experiments on images with a size of 512 $\times$ 512.

\noindent
\textbf{Spatial-Aware Style Encoder.}
As shown in Fig. \ref{fig2}, Spatial-Aware Style Encoder consists of two branches for receiving both style and spatial information. To fuse two different domains, we leverage SFT Layers in SFT-GAN \cite{wang2018recovering}. The detailed architecture of SFT Layer is shown in Fig. \ref{fig3} which does not use any normalization for all layers. 

\noindent
\textbf{Alpha Blender.}
Alpha Blender also follows the desing of Pix2PixHD but only downsampling three times and using three residual blocks. The detailed architecture is as follow: \\ $c7s1-32,d64,d128,d256,R256,R256,R256,u128,u64,\\u32-c7s1$ which uses IN for all layers.

\noindent
\textbf{Discriminator.}
Our design of discriminator also follows Pix2PixHD \cite{wang2018high} which utilize PatchGAN \cite{isola2017image}. We concatenate the masks and images as inputs to realize conditional GAN \cite{mirza2014conditional}. The detailed architecture is as follow: \\ $c64, c128, c256, c512$ which uses IN for all layers.

 \noindent
 \textbf{Training Details.}
 Our Dense Mapping Network and MaskVAE are both updated with the Adam optimizer \cite{kingma2014adam} ($\beta_{1} = 0.5$, $\beta_{2} = 0.999$, learning rate of $2e^{-4}$). For Editing Behavior Simulated Training, we reduce the learning rate to $5e^{-5}$. MaskVAE is trained with batch size of 16 and MaskGAN is trained with the batch size of 8.
\begin{figure}[t]
\begin{center}
 \includegraphics[width=0.7\linewidth]{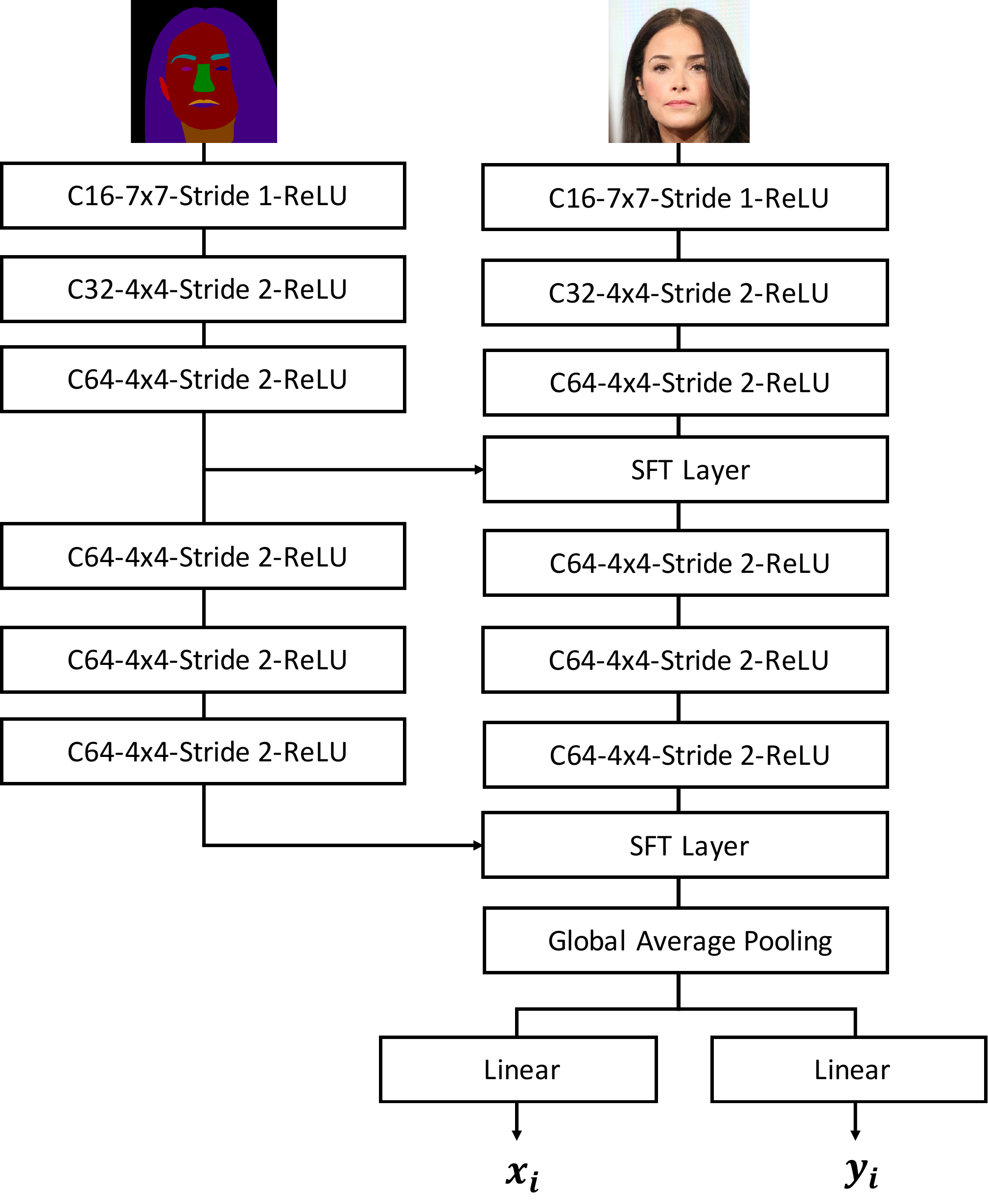}
\end{center}
   \vspace{-10pt}
   \caption{Architecture of Spatial-Aware Style Encoder.}
\label{fig:long}
\label{fig:onecol}
\label{fig2}
\end{figure} 

\begin{figure}[t]
\begin{center}
 \includegraphics[width=0.7\linewidth]{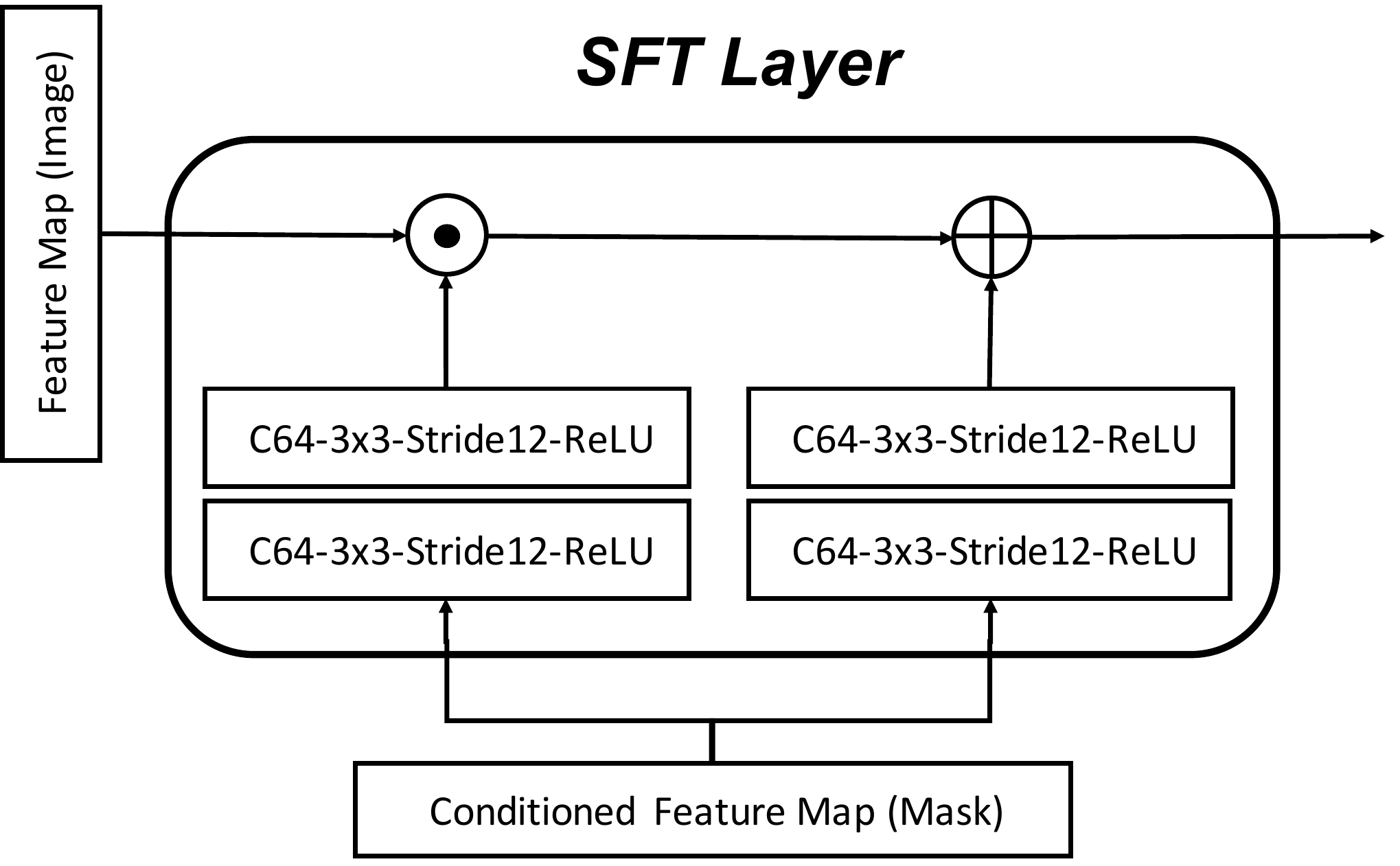}
\end{center} 
   \vspace{-10pt}
   \caption{Architecture of Spatial Feature Transform Layer.}
\label{fig:long}
\label{fig:onecol}
\label{fig3}
\end{figure} 
\begin{figure}[h]
\vspace{-6pt}
\begin{center}
 \includegraphics[width=0.8\linewidth]{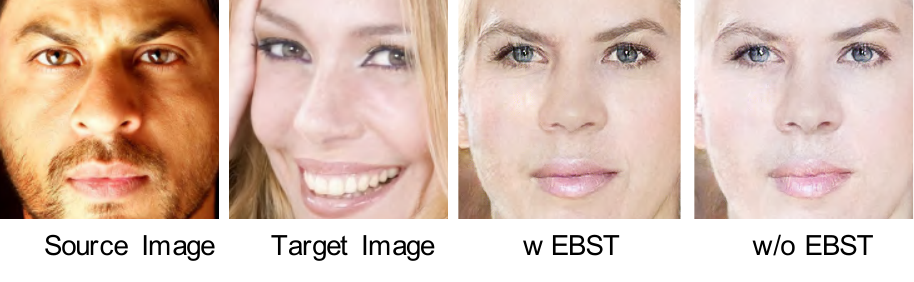}
\end{center}
   \vspace{-10pt}
   \caption{Visual comparisons of training with and without EBST.}
\label{fig:long}
\label{fig:onecol}
\label{fig4}
\end{figure} 
\begin{figure*}[t]
\begin{center}
 \includegraphics[width=1\linewidth]{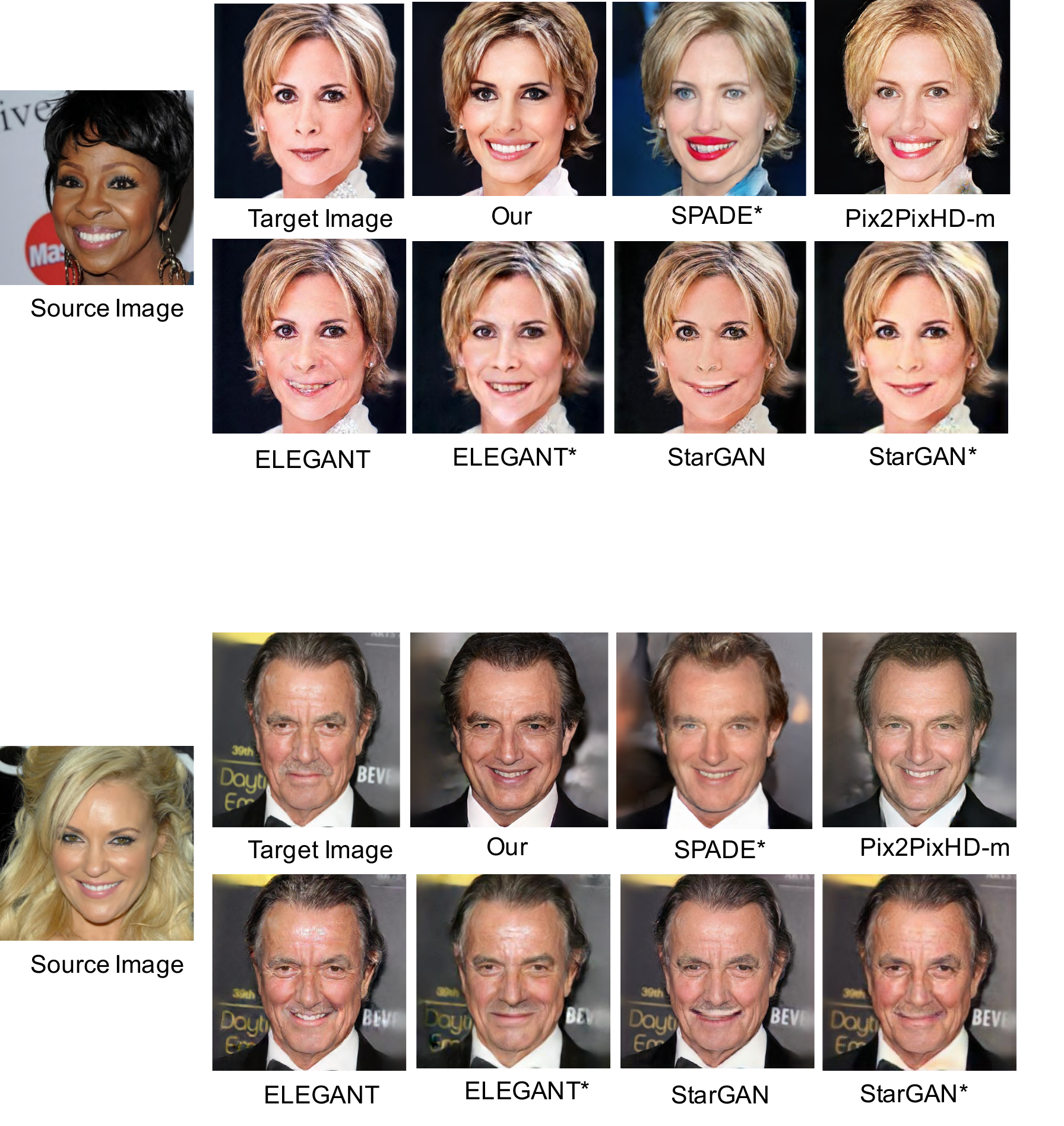}
\end{center}
   \caption{Visual results of attribute transfer for a specific attribute: \textbf{Smiling}. * means the model is trained with a size of 256 $\times$ 256.}
\label{fig:long}
\label{fig:onecol}
\label{fig5}
\end{figure*} 
\begin{figure*}[t]
\begin{center}
 \includegraphics[width=1\linewidth]{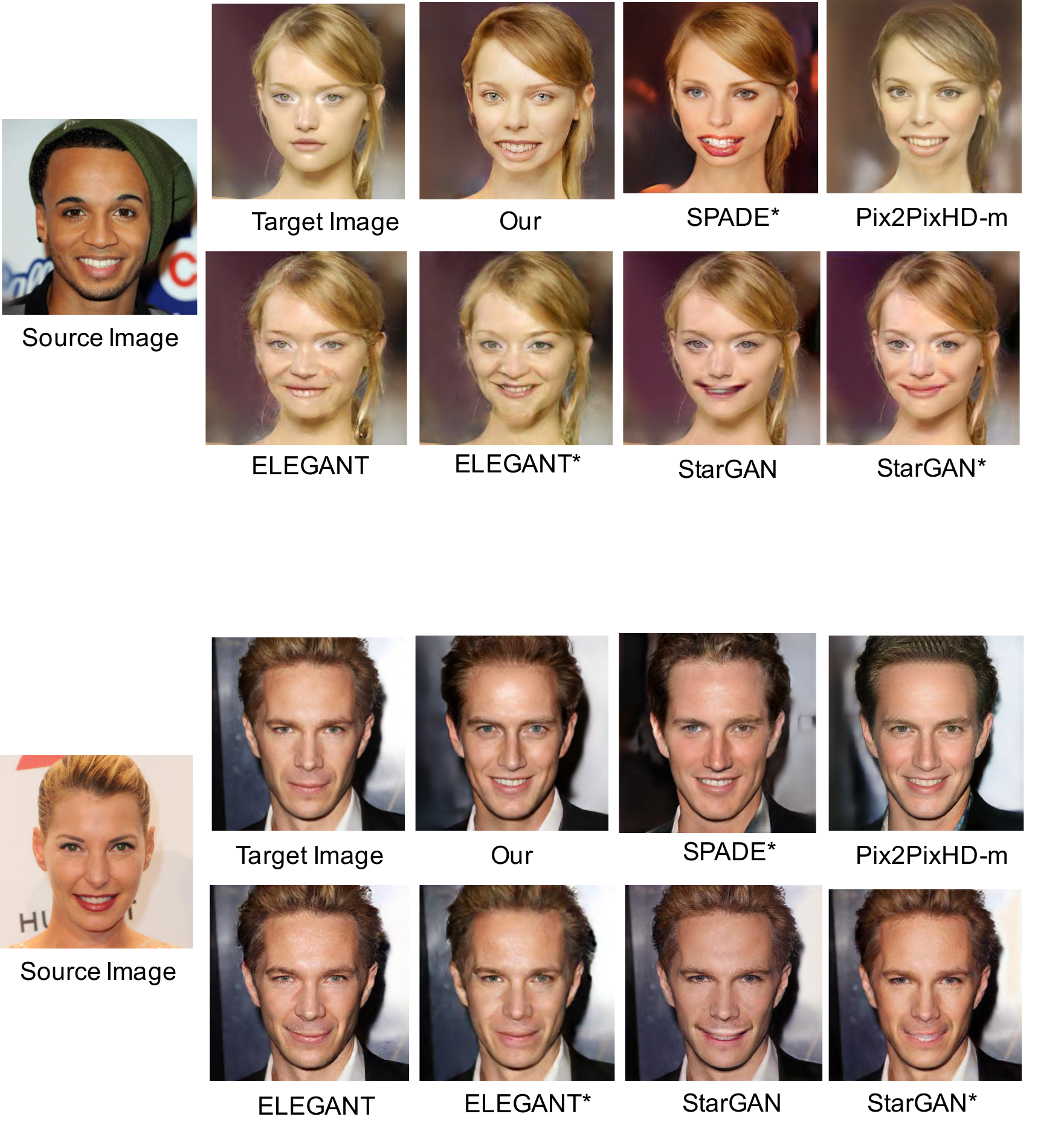}
\end{center}
   \caption{Visual results of attribute transfer for a specific attribute: \textbf{Smiling}. * means the model is trained with a size of 256 $\times$ 256.}
\label{fig:long}
\label{fig:onecol}
\label{fig6}
\end{figure*} 
\begin{figure*}[t]
\begin{center}
 \includegraphics[width=1\linewidth]{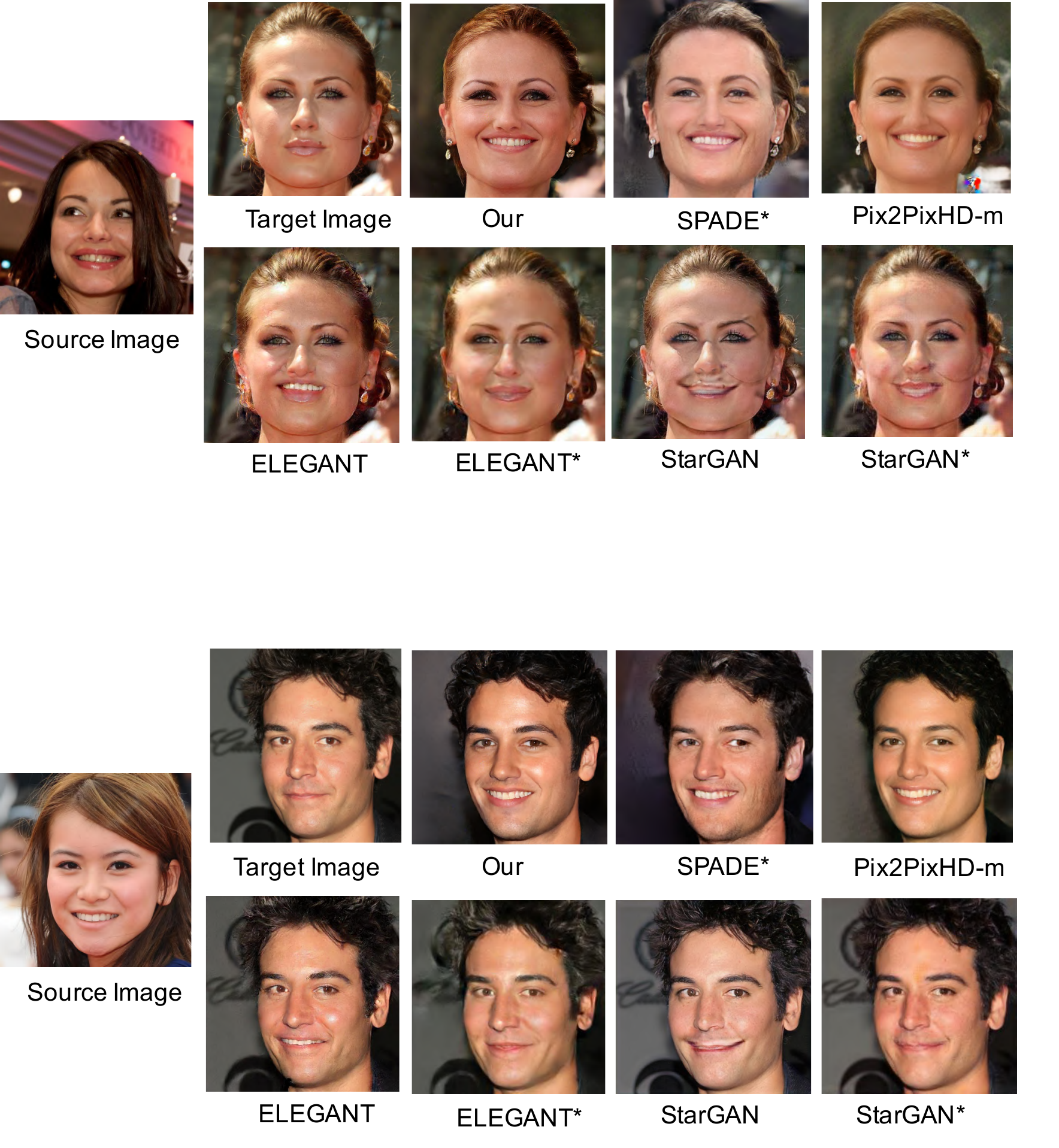}
\end{center}
   \caption{Visual results of attribute transfer for a specific attribute: \textbf{Smiling}. * means the model is trained with a size of 256 $\times$ 256.}
\label{fig:long}
\label{fig:onecol}
\label{fig7}
\end{figure*} 
\begin{figure*}[t]
\begin{center}
 \includegraphics[width=1\linewidth]{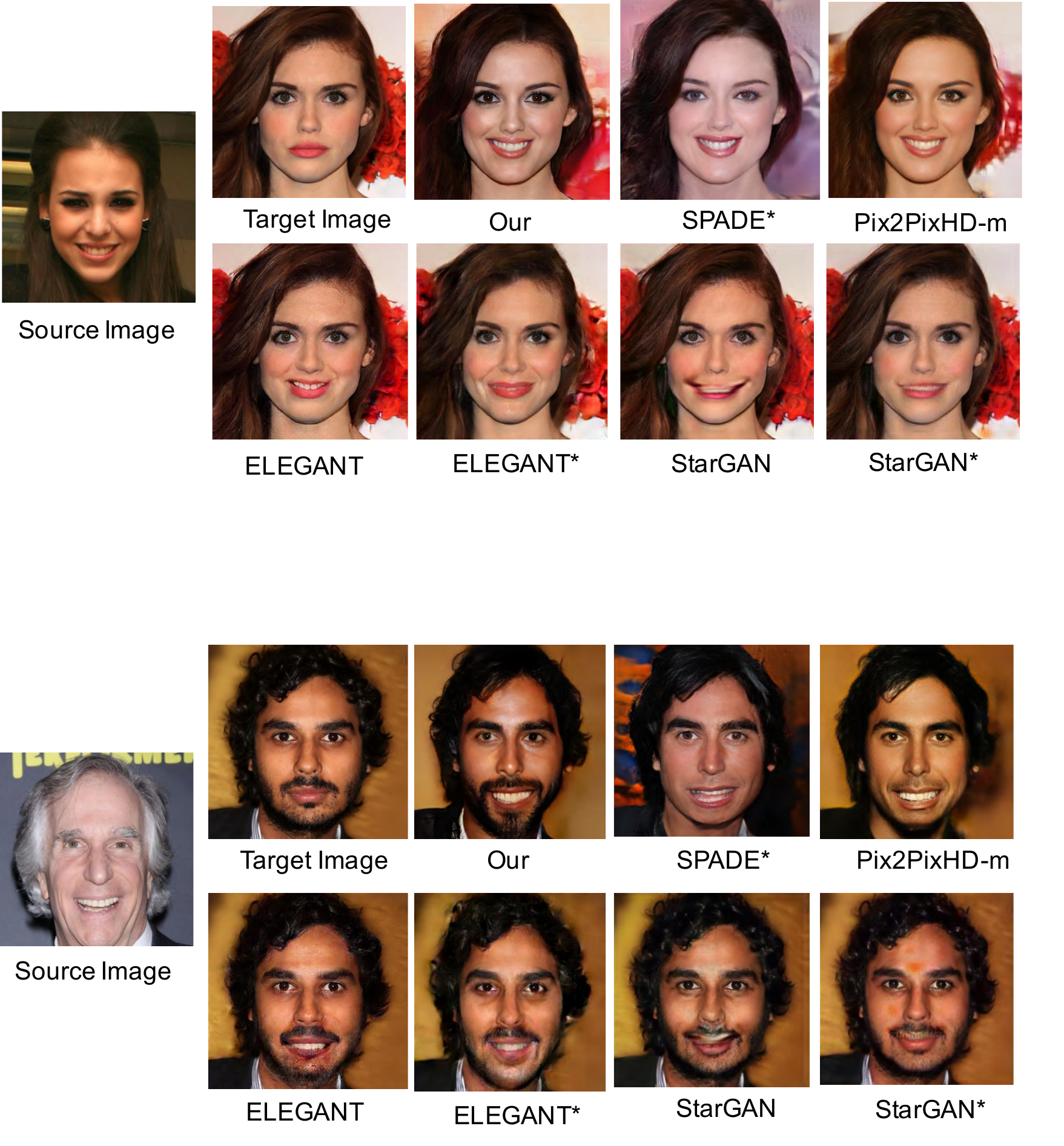}
\end{center}
   \caption{Visual results of attribute transfer for a specific attribute: \textbf{Smiling}. * means the model is trained with a size of 256 $\times$ 256.}
\label{fig:long}
\label{fig:onecol}
\label{fig8}
\end{figure*} 
\begin{figure*}[t]
\begin{center}
 \includegraphics[width=1\linewidth]{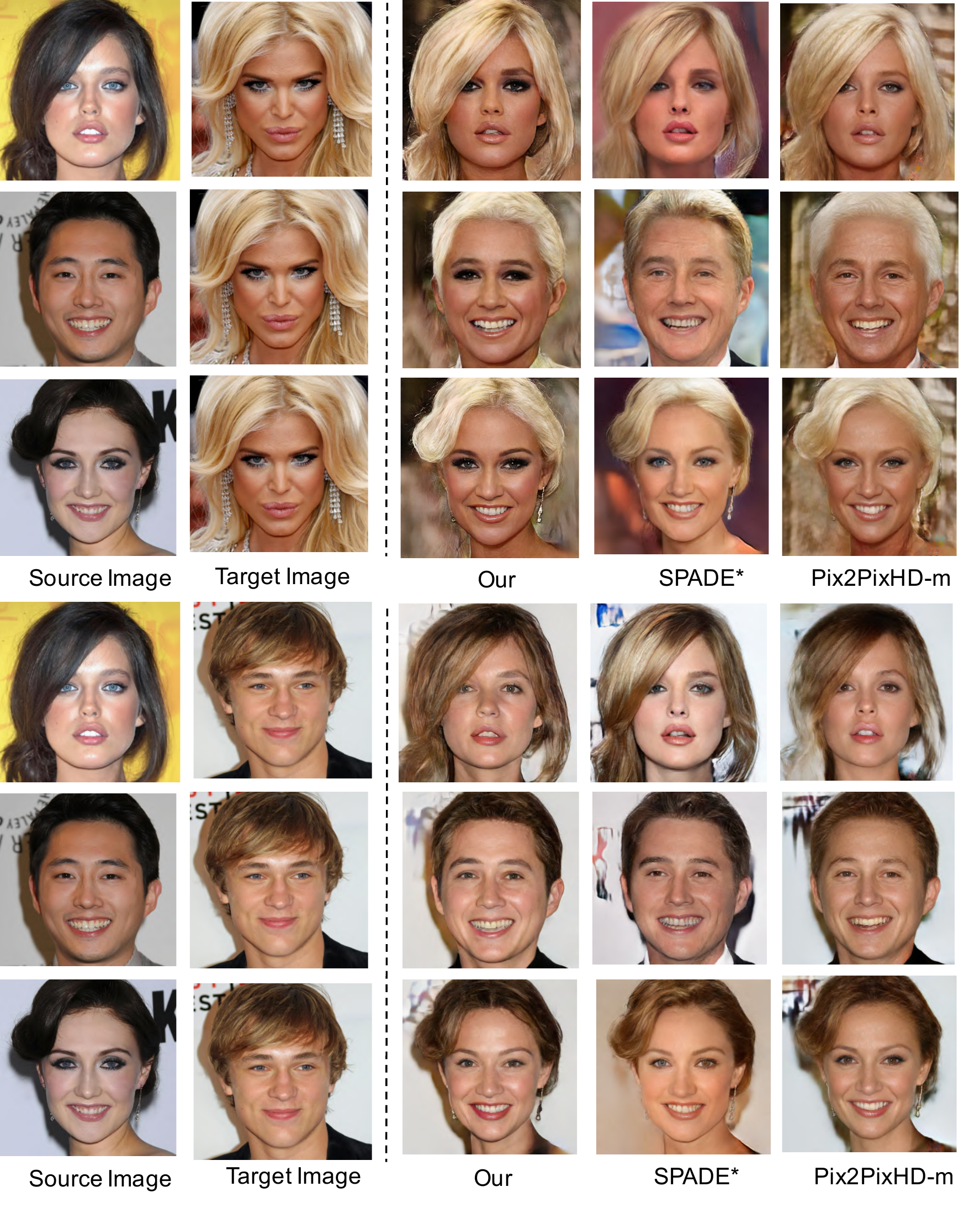}
\end{center}
   \caption{Visual results of style copy.}
\label{fig:long}
\label{fig:onecol}
\label{fig9}
\end{figure*} 
\begin{figure*}[t]
\begin{center}
 \includegraphics[width=1\linewidth]{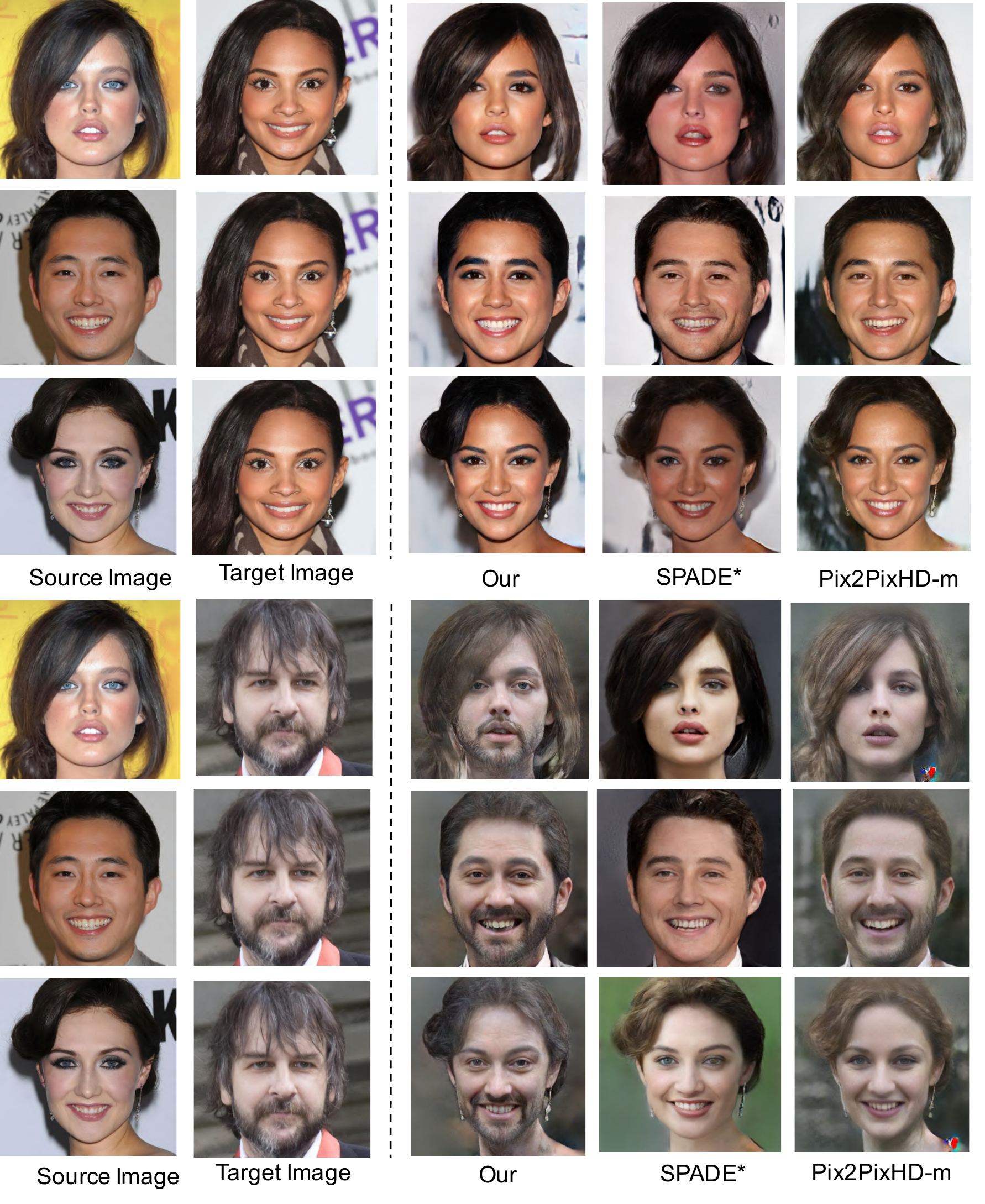}
\end{center}
   \caption{Visual results of style copy.}
\label{fig:long}
\label{fig:onecol}
\label{fig10}
\end{figure*} 
\begin{figure*}[t]
\begin{center}
 \includegraphics[width=1\linewidth]{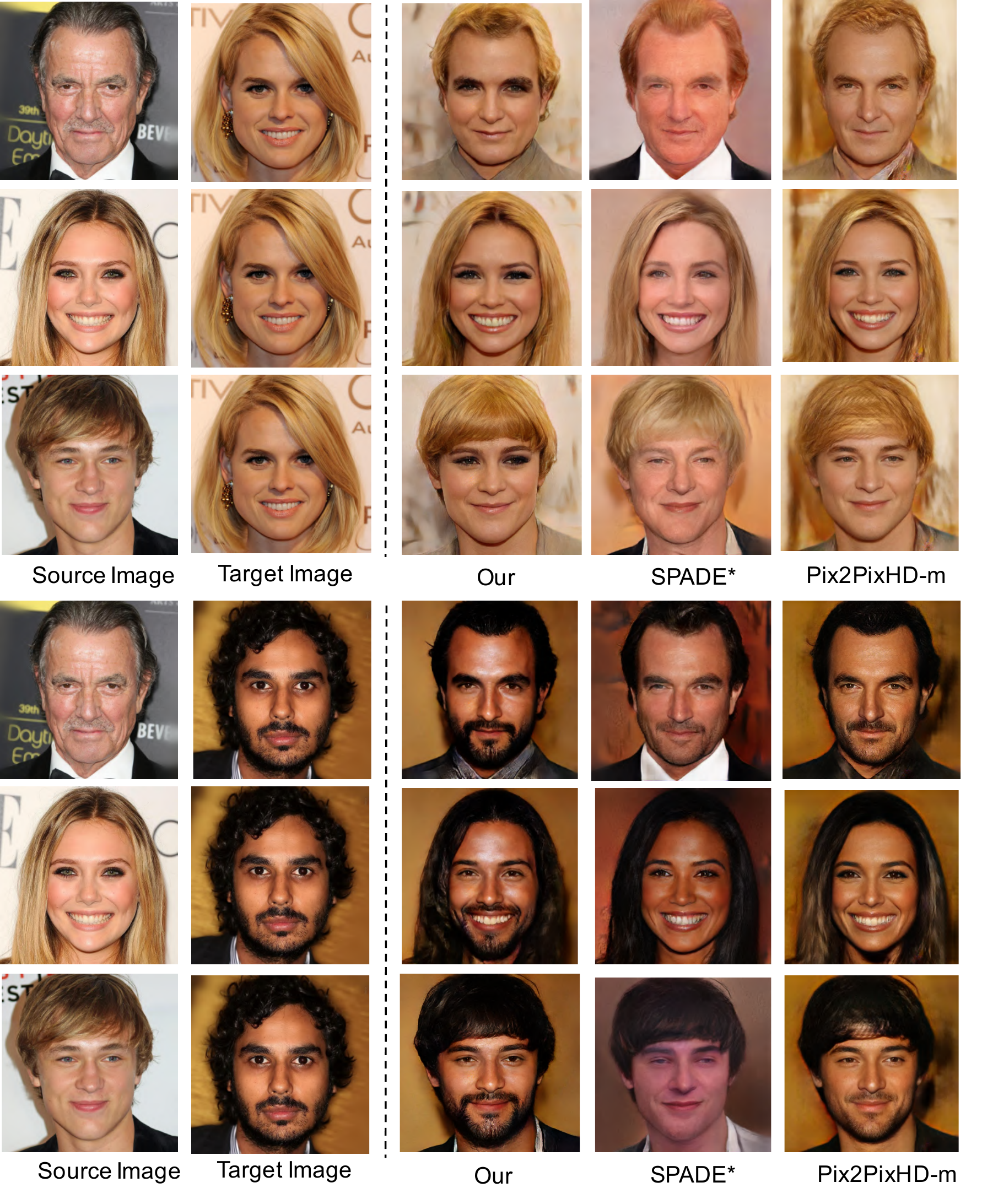}
\end{center}
   \caption{Visual results of style copy.}
\label{fig:long}
\label{fig:onecol}
\label{fig11}
\end{figure*} 
\begin{figure*}[t]
\begin{center}
 \includegraphics[width=1\linewidth]{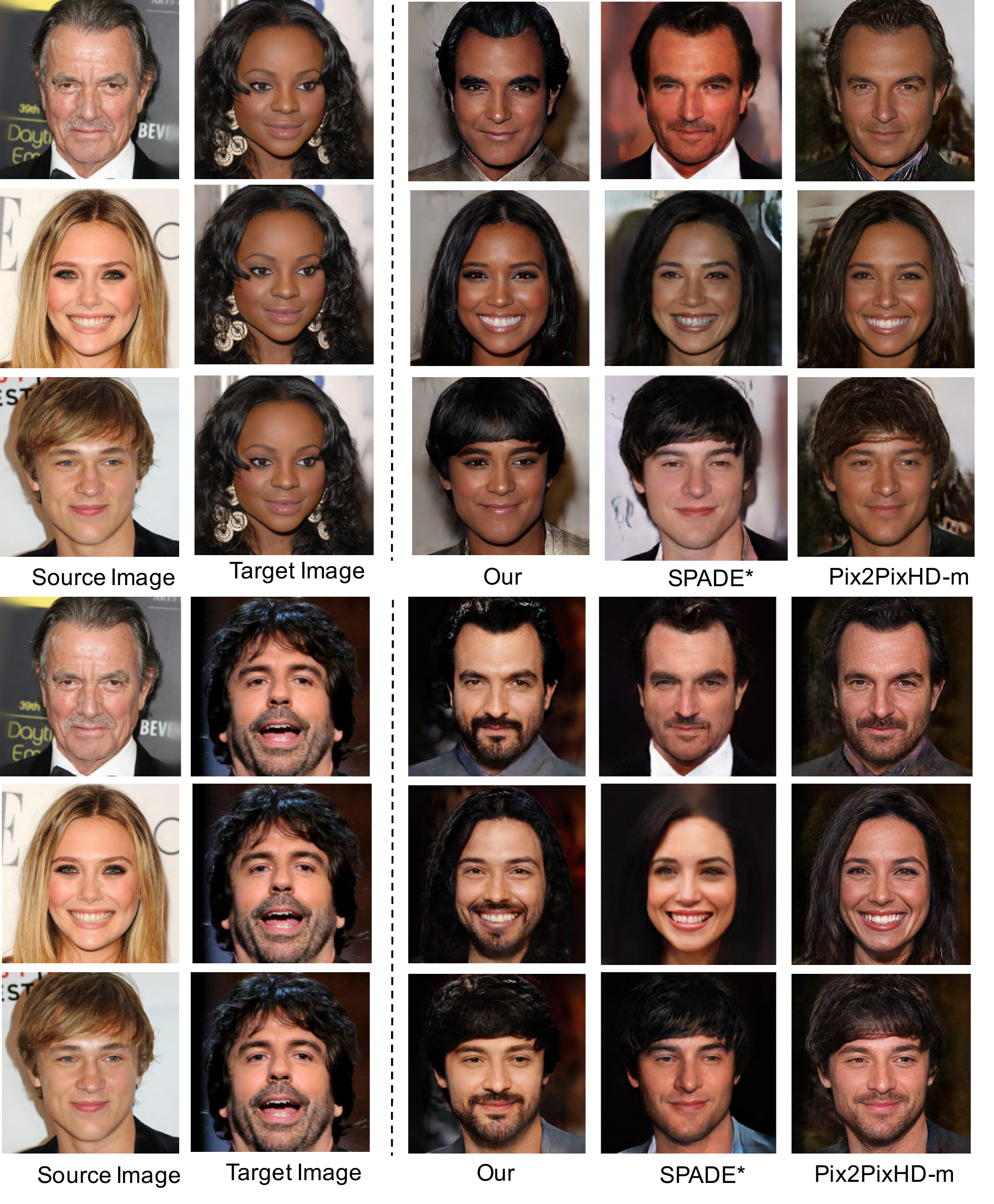}
\end{center}
   \caption{Visual results of style copy.}
\label{fig:long}
\label{fig:onecol}
\label{fig12}
\end{figure*} 
\section{Additional Ablation Study}
A simple quantitative comparison is shown in Table. \ref{tab1}. SFT layers utilize more parameters to fuse to different domains together. As a result, it is reasonable that SFT layers have better effect than concatenation.

In Fig. \ref{fig4}, we show a visual comparison of style copy. The results with EBST have better color saturation and attribute keeping quality (heavy makeup).

\section{Additional Visual Results}
In Fig. \ref{fig5}, Fig. \ref{fig6}, Fig. \ref{fig7}, and Fig. \ref{fig8}, we show additional visual results of attribute transfer for a specific attribute: \textbf{Smiling}. We compare our MaskGAN with state-of-the art methods including Pix2PixHD \cite{wang2018high} with modification, ELEGANT \cite{xiao2018elegant}, and StarGAN \cite{choi2018stargan}.

In Fig. \ref{fig9}, Fig. \ref{fig10}, Fig. \ref{fig11} and Fig. \ref{fig12}, we show additional visual results of style. We compare our MaskGAN with state-of-the art methods including Pix2PixHD \cite{wang2018high} with modification.

In the accompanying \href{https://www.youtube.com/watch?v=T1o38DFalWs}{video}, we demonstrate our interactive facial image manipulation interface. Users can edit the shape of facial components or add some accessories toward manipulating the semantic segmentation mask. 
\begin{table}
\scriptsize
\begin{center}
\begin{tabular}{c|c|c|c} 
\hline
Metric & Attribute cls. acc(\%) & Seg(\%) & FID\\ 
\hline
MaskGAN-concat & 63.1\hspace{0.5cm}61.3\hspace{0.5cm}84.8 & 90.8 & 27.13 \\ 
MaskGAN-SFT &\textcolor{red}{67.7}\hspace{0.5cm}\textcolor{red}{67.1}\hspace{0.5cm}\textcolor{red}{89.0}& \textcolor{red}{92.5} & \textcolor{red}{26.22}\\ 
\hline
GT & 96.9\hspace{0.5cm}88.1\hspace{0.5cm}95.4 & 93.4 & -\\
\hline
\end{tabular}
\end{center}
\vspace{-5pt}
\caption{Ablation study on style copy. Attribute types in attribute classification accuracy from left to right are \textbf{Male}, \textbf{Heavy Makeup}, and \textbf{No Beard}. P.S. The train/test split here is different from the main paper.} 
\label{tab1}
\end{table}

\end{document}